\documentclass[11pt]{article}

\newcommand{\bftab}{\fontseries{b}\selectfont}
\usepackage[a4paper]{geometry}
\usepackage[numbers,sort&compress]{natbib}
\usepackage{url}
\usepackage[dvipsnames]{xcolor}
\usepackage[
  colorlinks=true,
  linkcolor=blue!60!black,
  citecolor=RoyalBlue,
  urlcolor=RoyalBlue,
  filecolor=MidnightBlue,
  breaklinks=true,
  bookmarksopen=true,
  bookmarksnumbered=true,
  pdfstartview=FitH,
  pdfpagemode=UseOutlines
]{hyperref}

\usepackage{booktabs}
\usepackage{multirow}

\usepackage{graphicx}   
\usepackage{amsmath,amssymb}
\usepackage{pgffor}
\usepackage{subcaption}

\usepackage{fancyhdr}
\graphicspath{{Figures/}}

\usepackage{setspace}
\title{Whole-body CT attenuation and volume charts from routine clinical scans via evidence-grounded LLM report filtering} 

\author{
\hspace*{-1cm}%
\parbox{1.1\textwidth}{\center
Christian Wachinger$^{1,3,4,*}$, Bernhard Renger$^{1}$, Christopher Späth$^{1}$,\\
Jan Kirschke$^{2}$, Marcus Makowski$^{1}$\\
{\footnotesize $^{1}$Institute of Radiology, School of Medicine and Health,
Technical University of Munich, Munich, Germany}\\
{\footnotesize $^{2}$Institute of Neuroradiology, School of Medicine and Health,
Technical University of Munich, Munich, Germany}\\
{\footnotesize $^{3}$Munich Data Science Institute (MDSI), Technical University of Munich, Munich, Germany}\\
{\footnotesize $^{4}$Munich Center for Machine Learning (MCML), Munich, Germany}\\
{\footnotesize *Corresponding author: Christian Wachinger (christian.wachinger@tum.de)}
}
}
\date{}

\begin{document}

\newcommand{\longplotpathVol}{Figures/long_gamlss_vol_plot/}
\newcommand{\longplotpathInt}{Figures/long_gamm4_int_plot_col/}

\newcommand{\gamlsspathVol}{Figures/gamlss_plots_vol_filt2_col/}
\newcommand{\bootstrappathVol}{Figures/bootstrap_plots_vol/}
\newcommand{\derivativepathVol}{Figures/gamlss_plots_deriv_vol_filt2_col/}

\newcommand{\gamlsspathInt}{Figures/gamlss_plots_int_filt2_col/}
\newcommand{\bootstrappathInt}{Figures/bootstrap_plots_int/}
\newcommand{\derivativepathInt}{Figures/gamlss_plots_deriv_int_filt2_col/}

\def\organs{%
liver/liver,%
kidney_right/kidney right,%
spleen/spleen,%
lung_upper_lobe_right/lung upper lobe right,%
aorta/aorta,%
heart/heart,%
gluteus_minimus_right/gluteus minimus right,%
vertebrae_L3/vertebrae L3%
}

{\setstretch{1.45}

\maketitle

\begin{abstract}
Interpreting quantitative CT biomarkers, such as organ volume and tissue attenuation, requires large-scale healthy reference distributions. However, creating these is challenging because clinical datasets are often heavily enriched with pathology.
Here, we develop an evidence-grounded, cross-verified large language model (LLM) ensemble to filter pathological findings from radiology reports, enabling the construction of pathology-reduced cohorts from over 350,000 CT examinations. Five LLMs, first, flag structure-level abnormality candidates grounded in verbatim report evidence and, second, resolve disagreements via cross-verification. 
Using distribution-aware generalized additive models for location, scale, and shape, we establish comprehensive whole-body reference charts for 106 anatomical structures (volumes and attenuation) across adulthood, accounting for age, sex, contrast enhancement, and acquisition parameters. Longitudinal analyses reveal structure- and contrast-dependent changes distinct from cross-sectional trends. These resources facilitate covariate-adjusted centile scoring from routine CT, supporting standardized quantitative phenotyping, multi-site imaging studies, and scalable opportunistic screening research.
\end{abstract}
}

\section{Introduction}
Routine computed tomography (CT) contains a wealth of quantitative information about anatomy and tissue composition, but has historically been difficult to extract at scale.
Advances in deep learning now enable accurate, automated segmentation for more than 100 anatomical structures across heterogeneous clinical CT acquisitions and to extract quantitative biomarkers such as organ volumes and mean attenuation \citep{wasserthalTotalSegmentatorRobustSegmentation2023}.
Unlike MRI intensities, CT attenuation is expressed on a standardized physical scale (Hounsfield Units, HU), facilitating its use as an opportunistic marker for tissue composition and disease processes. 
Examples include hepatic steatosis and liver fat assessment~\citep{boyceHepaticSteatosisFatty2010,starekovaQuantificationLiverFat2021}, pancreatic fatty infiltration~\citep{tanabeAutomatedWholevolumeMeasurement2023,bhallaAssociationPancreaticFatty2022}, and opportunistic screening for osteoporosis and sarcopenia based on vertebral and muscle attenuation~\citep{jangOpportunisticOsteoporosisScreening2019,boutinValueAddedOpportunisticCT2020,pickhardtValueaddedOpportunisticCT2022,pickhardtOpportunisticScreeningRadiology2023}.

A central limitation of raw CT-derived biomarkers is that they are difficult to interpret in a single examination. Organ volumes and HU values vary systematically with age and sex and are further influenced by acquisition factors, most prominently intravenous contrast and scanner characteristics. Normative modeling addresses this by quantifying expected biological and technical variation and expressing an individual measurement as a standardized deviation from an age- and covariate-matched reference distribution, thereby flagging atypical values \cite{cole1990lms}. This reference-chart paradigm is well established in medicine through pediatric growth charts \citep{coleDevelopmentGrowthReferences2012} and has more recently been extended to brain MRI markers \citep{bethlehemBrainChartsHuman2022,rutherfordNormativeModelingFramework2022,wachingerBodyChartsCT2025}. 
However, whole-body CT reference modeling requires large cohorts to capture nonlinear age trajectories and to estimate distribution tails reliably enough for stable extreme centiles, even with heterogeneous contrast and scanner conditions.

Whole-body CT cannot be acquired at population scale in healthy volunteers because it involves ionizing radiation. As a result, cohorts comparable to population biobanks with whole-body MRI, such as UKB \cite{littlejohnsUKBiobankImaging2020} or GNC \cite{bambergWholeBodyMRImaging2015}, are not available for CT, and the most abundant source of data is the clinical picture archiving and communication system (PACS). PACS cohorts offer unmatched scale, anatomical breadth, and longitudinal follow-up, but they are enriched for pathology and incidental findings. This creates a core challenge for CT reference modeling: if examinations with relevant abnormalities remain in the modeling pool, disease-related changes can inflate variability, bias distribution tails, and shift centiles, weakening precisely the extreme scores that matter most for interpretation. 
Radiology reports are the most scalable way to reduce this pathology burden because they document the expert clinical interpretation of the scan \cite{esrGoodPracticeRadiological2011}. 
Yet reports are unstructured and heterogeneous in style, which makes structure-level curation difficult with rules alone and infeasible by manual review at the scale required for reference estimation \cite{cai2016natural}. 
Recent open-weight large language models (LLMs) can support report parsing across institutions and languages, but most medical applications have focused on text generation, question answering, and documentation support, rather than pathology identification for cohort construction \cite{zhou2023survey,woo2025use}.

Here, we address this barrier with a report-driven cohort curation framework that targets large-scale normative modeling at the scale of $350{,}000$ routine CT examinations.
We introduce a cross-verified, evidence-grounded filtering pipeline in which a multi-model ensemble extracts structure-level abnormality candidates from radiology reports, links each candidate to verbatim supporting sentences, and an independent verification stage adjudicates disputed candidates using pooled evidence snippets. To improve robustness against model-specific failure modes, we use five open-weight LLMs spanning distinct model lineages and medical domain adaptation. 

\begin{figure*}[t]
  \centering
  \includegraphics[width=\textwidth]{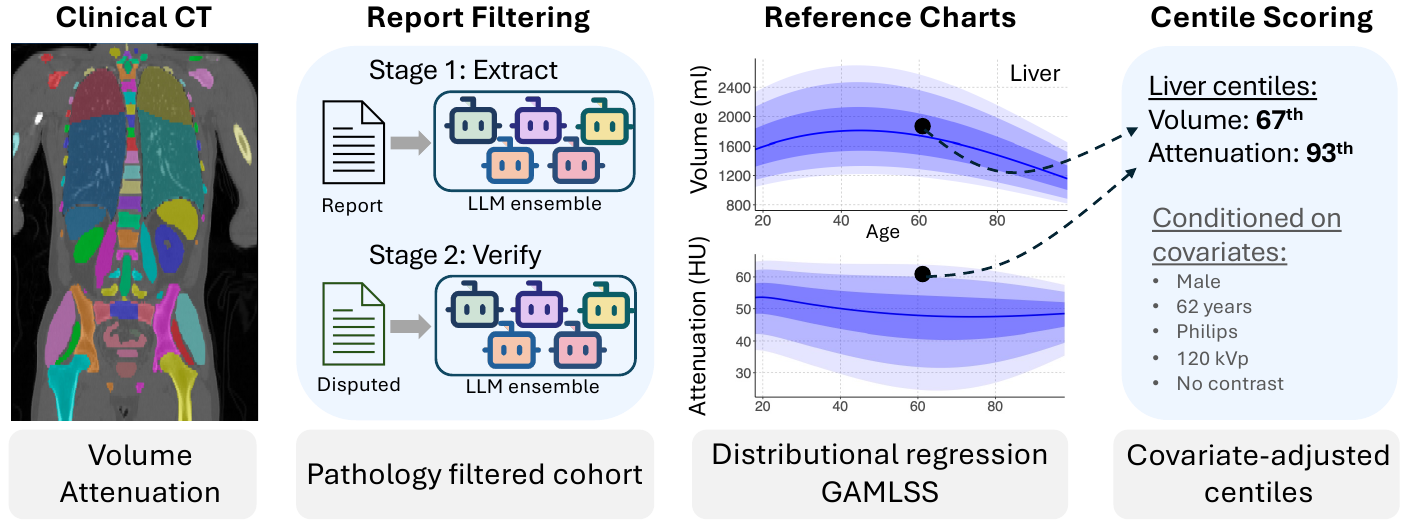}
  \caption{\textbf{Study overview.} Routine clinical CT examinations are automatically segmented to derive structure-wise volumes and mean attenuation (HU) alongside acquisition covariates. Radiology reports are parsed with a cross-verified, evidence-grounded LLM ensemble: five models propose abnormal structures with verbatim supporting sentences (Stage~1), and disputed candidates are adjudicated by a verification step using pooled evidence snippets (Stage~2), yielding a pathology-filtered cohort for reference modeling. Distribution-aware reference charts are then estimated with GAMLSS for volumes and attenuation, conditioning on biological and acquisition covariates. For a new examination, the same feature extraction maps measurements to covariate-adjusted centile scores. Icons: Lucide (ISC license).}
  \label{fig:overview}
\end{figure*}

Building on this curated foundation, we establish whole-body reference charts for structure-wise CT attenuation (HU) across adulthood and, as a secondary contribution, update our previously published organ-volume charts \citep{wachingerBodyChartsCT2025}. The key advance relative to prior work is that pathology-filtered cohort construction is performed at scale using evidence-grounded, cross-verified report parsing, enabling distributional (not only mean) reference modeling of both volumes and attenuation. 
We estimate cross-sectional reference distributions with generalized additive models for location, scale, and shape (GAMLSS) to capture nonlinear age effects and non-Gaussian biomarker distributions; for attenuation, we additionally stratify by contrast status to account for systematic acquisition-related shifts in HU \citep{rigbyGeneralizedAdditiveModels2005,borghiConstructionWorldHealth2006}.
Finally, leveraging repeated imaging in clinical practice, we quantify within-subject change longitudinally and contrast longitudinal change with cross-sectional age trends.
Figure~\ref{fig:overview} summarizes the end-to-end workflow from report-based pathology filtering to covariate-adjusted centile scoring, enabling per-structure benchmarking of routine clinical scans across heterogeneous real-world protocols.

\newpage

\begin{figure}
  \centering
  \includegraphics[width=0.8\textwidth]{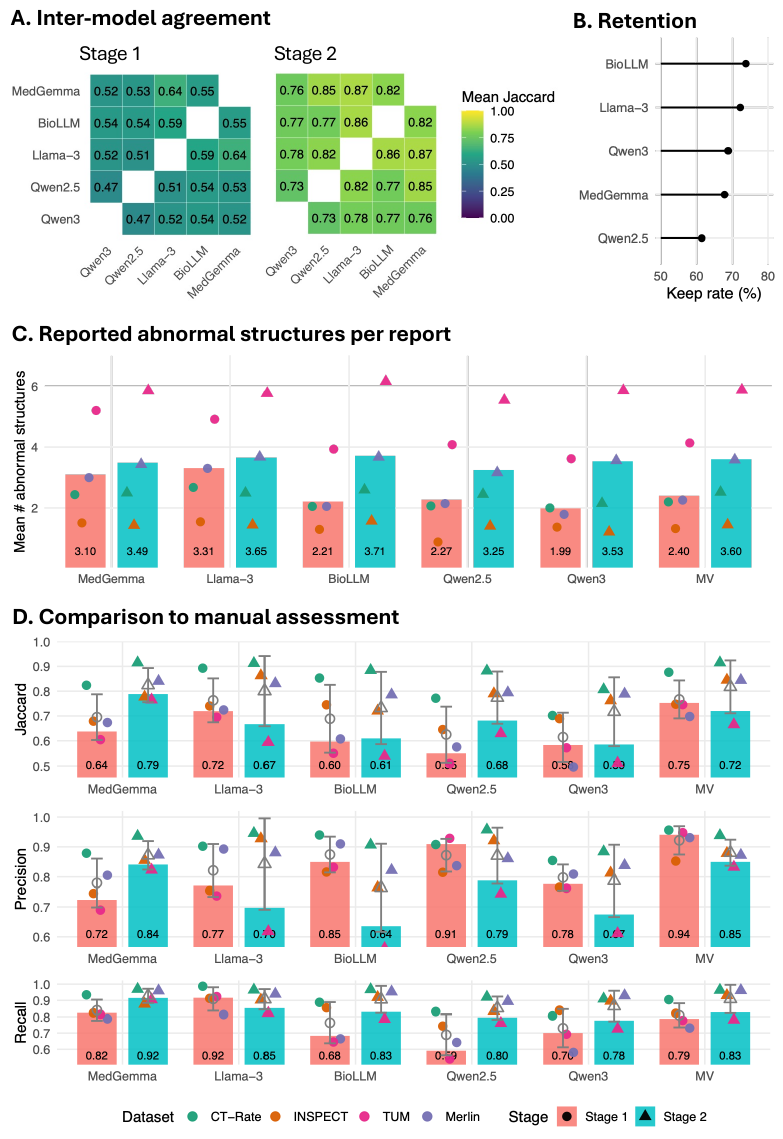}
\caption{\label{fig:llm_filtering_summary}
Summary of two-stage LLM-based report filtering and validation against manual structure labels.
\textbf{(A)} Inter-model consistency measured as mean pairwise Jaccard overlap between abnormal-structure sets across the five LLMs, shown for Stage~1 and Stage~2.
\textbf{(B)} Verification stringency on disputed candidates: percentage of Stage~1 disputed structure flags retained as abnormal by each verifier model. 
\textbf{(C)} Mean number of abnormal anatomical structures flagged per report for each method (five LLMs  and majority vote, MV) and both stages. 
\textbf{(D)} Agreement with manual structure labels: Jaccard, precision, and recall for both stages. 
 Bars show dataset-size--weighted global projected performance (from the full corpus). Colored points show dataset-specific projected performance; gray open markers and error bars indicate the unweighted mean $\pm$ SD across datasets (Stage~1: circle; Stage~2: triangle).
}
\end{figure}

\section{Results}

\subsection{LLM-based reference cohort construction from clinical CT}
To enable whole-body CT reference modeling, we assembled a multi-source CT dataset from our TUM PACS (292{,}914 examinations), CT-Rate (19{,}399) \cite{hamamciDevelopingGeneralistFoundation2025}, INSPECT (13{,}677) \cite{huangINSPECTMultimodalDataset2023}, Merlin (24{,}716) \cite{blankemeierMerlinVisionLanguage2024}, and TotalSegmentator (1{,}209)  \cite{wasserthalTotalSegmentatorRobustSegmentation2023}, yielding 351{,}915 examinations (see Methods \ref{sec:cohort}). 
{This aggregation was intentional to capture the biological, technical, and clinical heterogeneity of routine CT while providing sufficient sample size for distributional reference modeling across many structures.}
All scans were processed with TotalSegmentator to segment 106 structures and underwent automated quality control (see Methods \ref{sec:image}). 
Since these scans were acquired in clinical routine, residual pathology can inflate variability and bias distribution tails. 
We therefore developed a cross-verified, evidence-grounded LLM pipeline to identify structure-level abnormalities from radiology reports and adjudicate disputed candidates. 
{Report filtering was performed on 39 broader anatomical targets, derived from the 106 segmented structures by merging related structures.}
Five open-weight LLMs first extract abnormal structures independently (Stage~1); candidates without cross-model consensus are then pooled and re-evaluated in a verification stage (Stage~2), see Methods \ref{sec:llmFilt}.

Fig.~\ref{fig:llm_filtering_summary} summarizes the filtering results, where Stage~2 reduced model dependence while improving agreement with manual annotations. 
Cross-verification markedly increased inter-model consistency (Fig.~\ref{fig:llm_filtering_summary}A). Mean pairwise Jaccard overlap across the five LLMs rose from 0.541 in Stage~1 to 0.803 in Stage~2 ($+0.262$), indicating substantially reduced dependence on the specific model used. Exact set agreement likewise increased: the proportion of reports for which all five models returned identical abnormal-structure sets increased from 9.53\% to 30.07\% ($+20.54$ percentage points). 
From the pooled candidates, the verification keeps 61.4--73.8\% as abnormal, dropping 26.2--38.6\% (Fig.~\ref{fig:llm_filtering_summary}B). 

The mean number of abnormal structures flagged per report increased from Stage~1 to Stage~2 for all methods (Fig.~\ref{fig:llm_filtering_summary}C).
In Stage~1, prevalence varied substantially across single-model outputs (overall means 1.99--3.31), whereas Stage~2 reduced this dispersion (3.25--3.71), consistent with the verification step harmonizing disputed candidates across models. 
The magnitude of the increase from Stage~1 to Stage~2 differed markedly by method: BioLLM and Qwen3 showed the largest increases. Consistent with its design, majority voting (MV) yields a mid-range prevalence.

Fig.~\ref{fig:llm_filtering_summary}D illustrates the agreement with manual structure labels across four datasets. 
In unweighted per-dataset summaries, Stage~2 improved set overlap (Jaccard) for all methods relative to Stage~1. In the dataset-size--weighted aggregate, MedGemma showed the strongest gain in Jaccard overlap from Stage~1 to Stage~2 ($+0.15$), yielding the highest overall overlap among all approaches; with the exception of Llama, other methods also improved in Stage~2. Notably, majority voting over Stage~1 outputs achieved higher Jaccard overlap than any individual Stage~1 model, whereas the Stage~2 majority vote was slightly lower than Stage~1. Importantly, MedGemma in Stage~2 exceeded the initial majority voting in Jaccard overlap, motivating its selection as the final filtering strategy. Performance varied systematically by dataset: CT-Rate consistently achieved the highest Jaccard overlap across models and stages, whereas PACS was almost always the lowest-performing dataset, indicating that report characteristics beyond model choice contribute substantially to achievable filtering accuracy.

Because the primary application is pathology filtering prior to reference modeling, recall is particularly important: 
failing to identify abnormal structures allows pathology to remain in the modeling pool, which can bias the resulting reference trajectories.
{We therefore compared the final cross-verified strategy against the strongest simpler alternative, i.e., single-LLM filtering with Llama Stage~1.
In the weighted aggregate, MedGemma Stage~2 and Llama Stage~1 both achieved high recall ($0.92$), but MedGemma Stage~2 achieved this sensitivity with clearly higher precision, thereby preserving the ability to remove report-described pathology while reducing unnecessary exclusion of structures. 
In addition, MedGemma Stage~2 exceeded both individual Stage~1 models and the Stage~1 majority vote in weighted Jaccard overlap. 
Additional region-stratified target-level validation showed anatomical variation in filtering performance, with MedGemma Stage~2 maintaining high recall across regions (0.845--0.943) and outperforming Llama Stage~1 in regional recall for all anatomical groups (Supplement Section S2.2).
Together, these findings show that the cross-verified pipeline improved the precision--recall trade-off and reduced dependence on the arbitrary choice of a single LLM; accordingly, MedGemma Stage~2 was used for downstream cohort construction.}

\begin{figure*}[p]
\centering

\setlength{\tabcolsep}{0pt}
\setlength{\fboxsep}{0pt}

\begin{minipage}{0.24\textwidth}
\centering
\footnotesize liver\\[1ex]
\includegraphics[width=\linewidth]{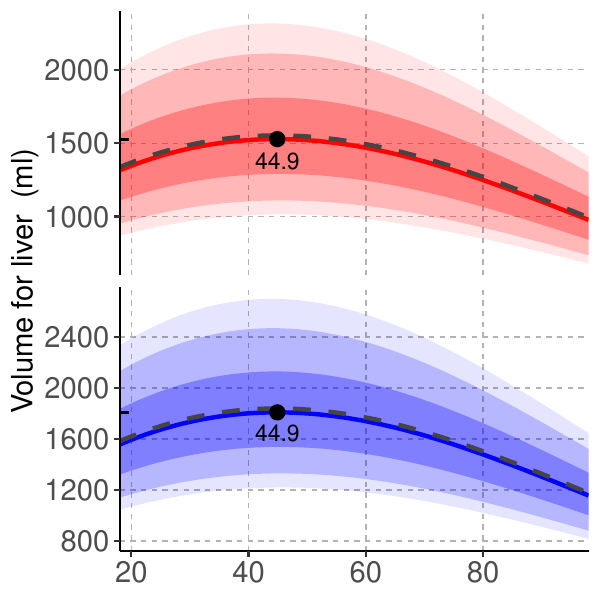}\\[1ex]
\includegraphics[width=\linewidth]{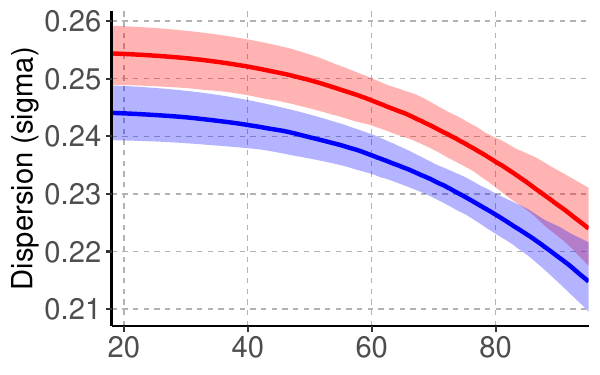}\\[1ex]
\includegraphics[width=\linewidth]{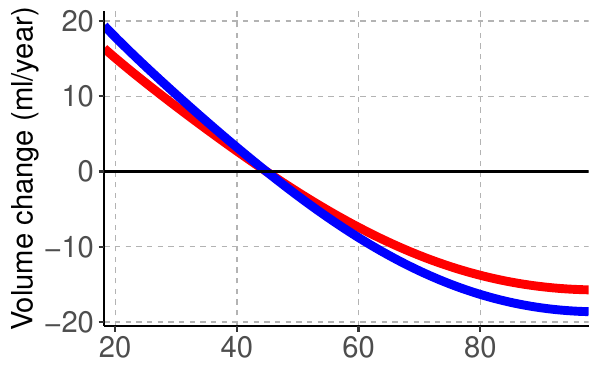}
\end{minipage}\hfill
\begin{minipage}{0.24\textwidth}
\centering
\footnotesize kidney right\\[1ex]
\includegraphics[width=\linewidth]{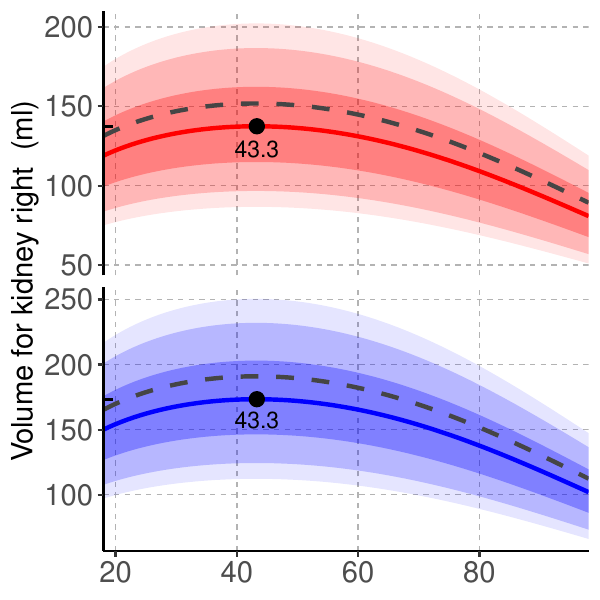}\\[1ex]
\includegraphics[width=\linewidth]{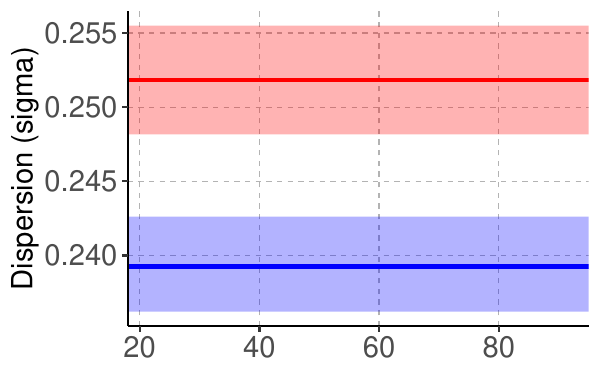}\\[1ex]
\includegraphics[width=\linewidth]{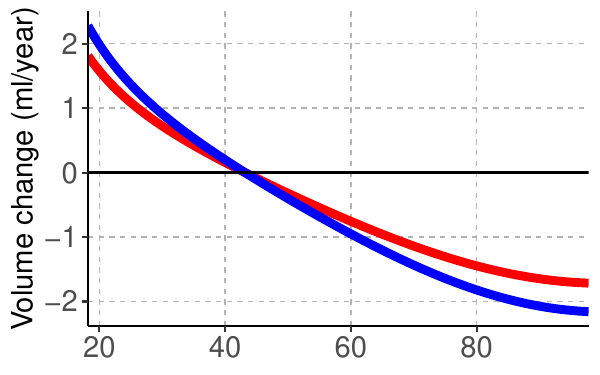}
\end{minipage}\hfill
\begin{minipage}{0.24\textwidth}
\centering
\footnotesize spleen\\[1ex]
\includegraphics[width=\linewidth]{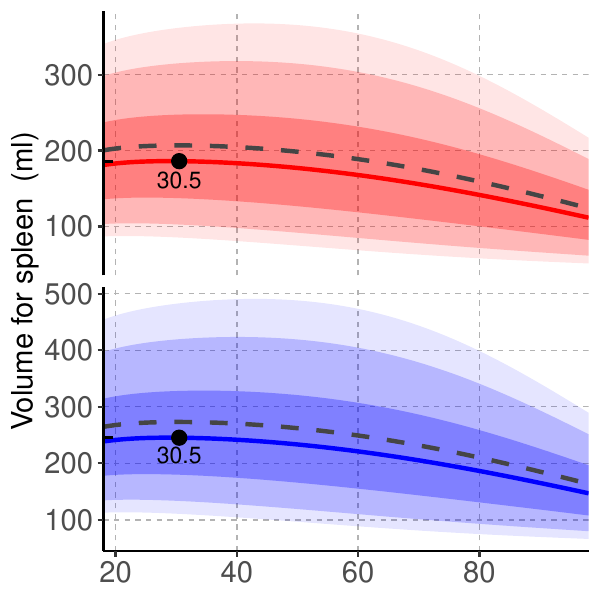}\\[1ex]
\includegraphics[width=\linewidth]{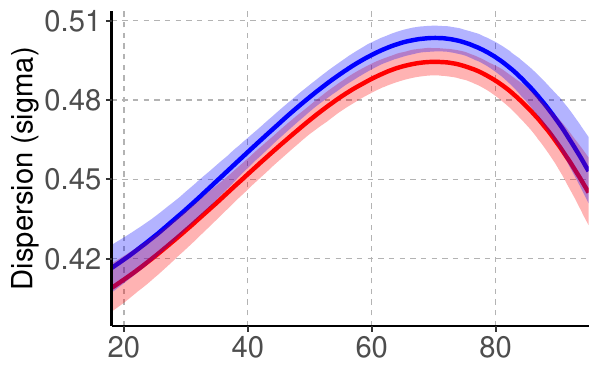}\\[1ex]
\includegraphics[width=\linewidth]{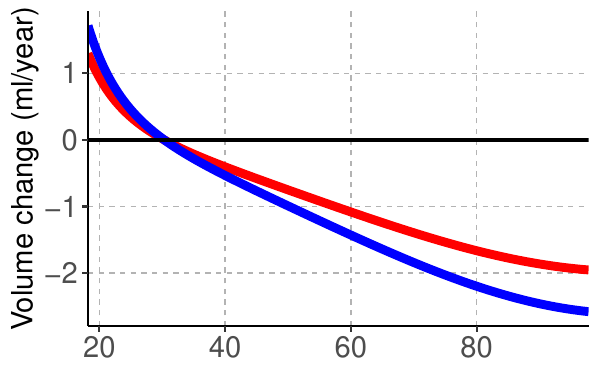}
\end{minipage}\hfill
\begin{minipage}{0.24\textwidth}
\centering
\footnotesize lung upper lobe right\\[1ex]
\includegraphics[width=\linewidth]{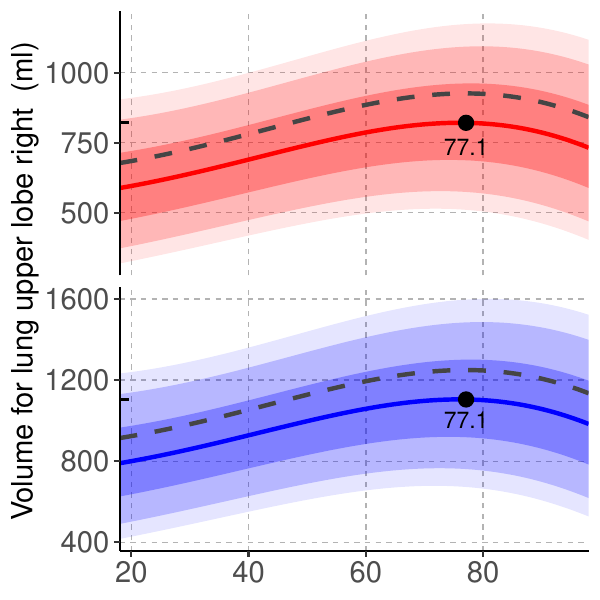}\\[1ex]
\includegraphics[width=\linewidth]{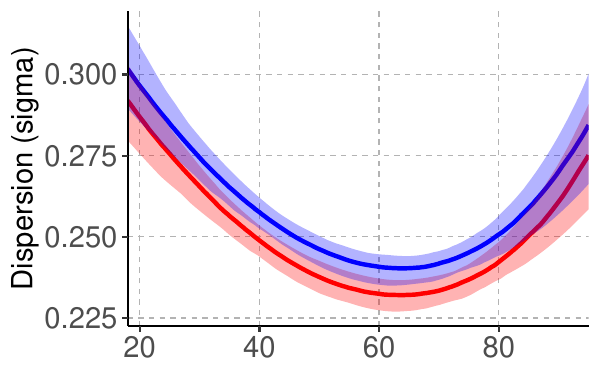}\\[1ex]
\includegraphics[width=\linewidth]{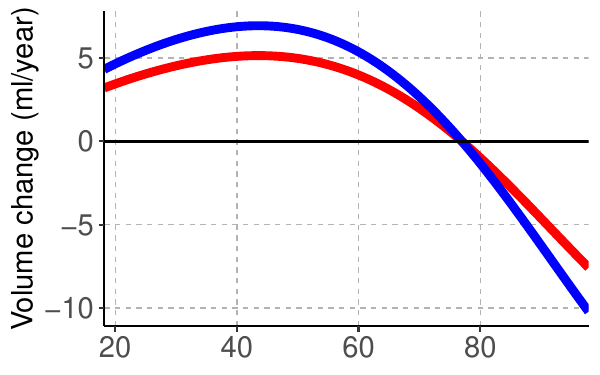}
\end{minipage}

\medskip

\begin{minipage}{0.24\textwidth}
\centering
\footnotesize aorta\\[1ex]
\includegraphics[width=\linewidth]{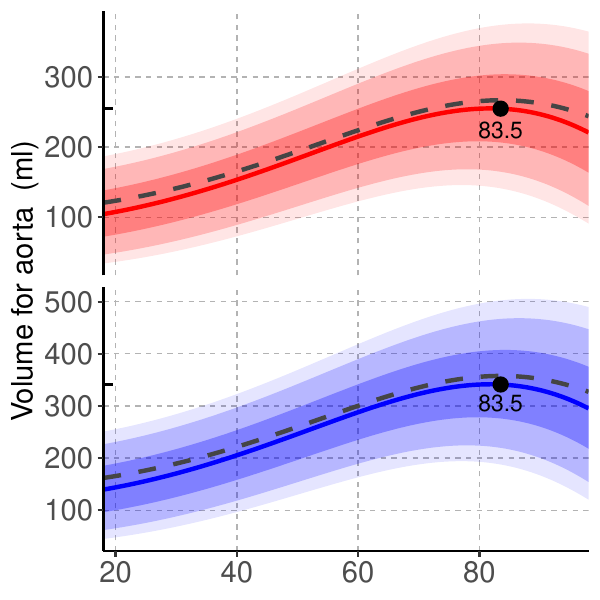}\\[1ex]
\includegraphics[width=\linewidth]{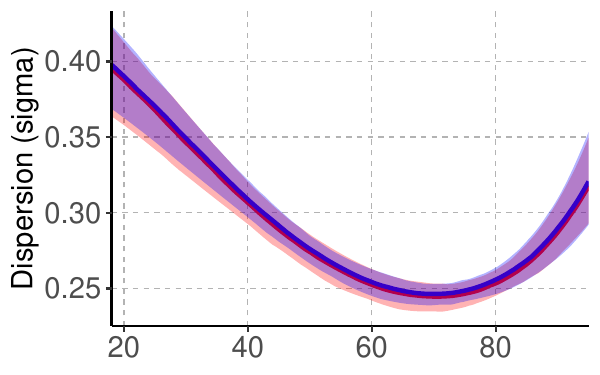}\\[1ex]
\includegraphics[width=\linewidth]{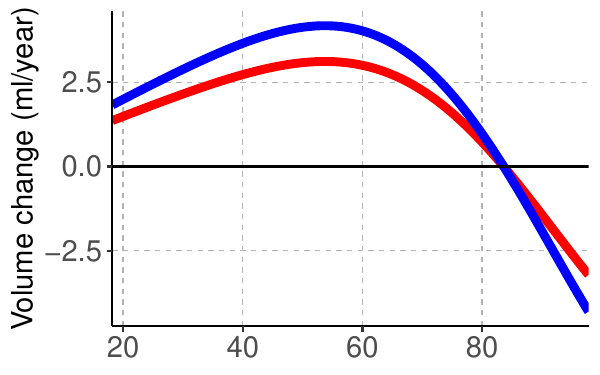}
\end{minipage}\hfill
\begin{minipage}{0.24\textwidth}
\centering
\footnotesize heart\\[1ex]
\includegraphics[width=\linewidth]{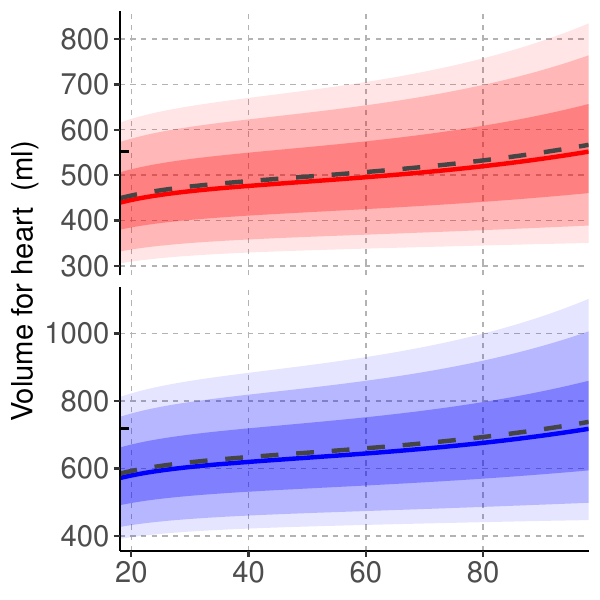}\\[1ex]
\includegraphics[width=\linewidth]{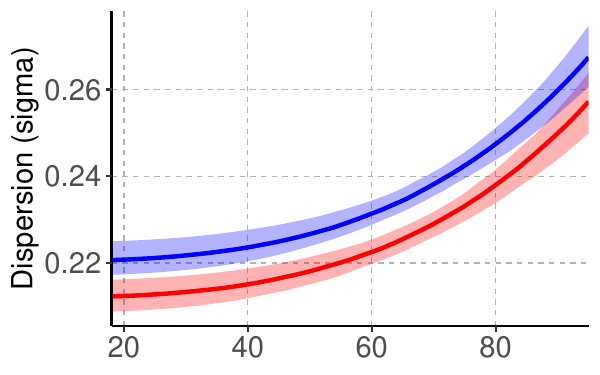}\\[1ex]
\includegraphics[width=\linewidth]{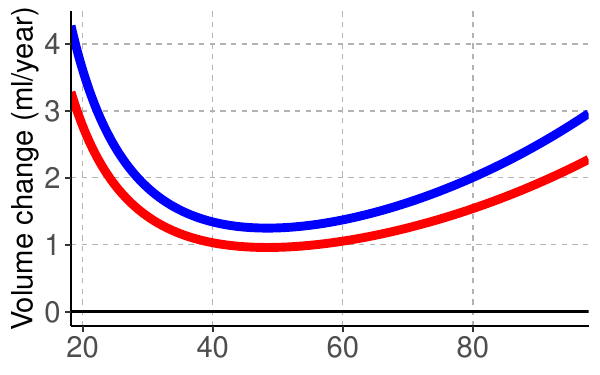}
\end{minipage}\hfill
\begin{minipage}{0.24\textwidth}
\centering
\footnotesize gluteus minimus right\\[1ex]
\includegraphics[width=\linewidth]{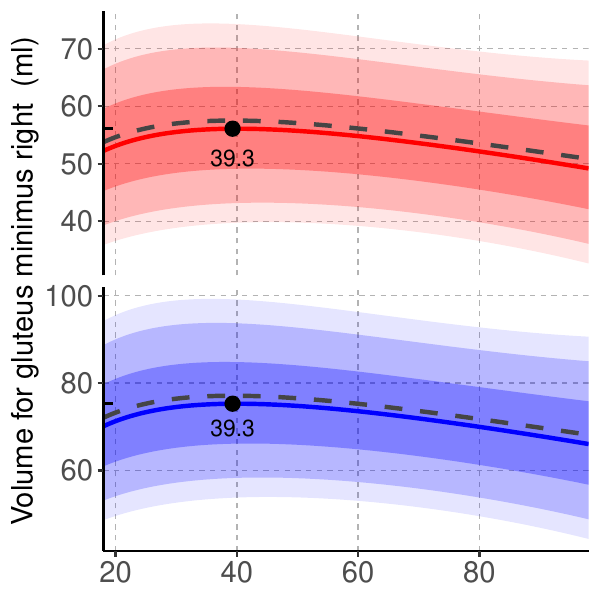}\\[1ex]
\includegraphics[width=\linewidth]{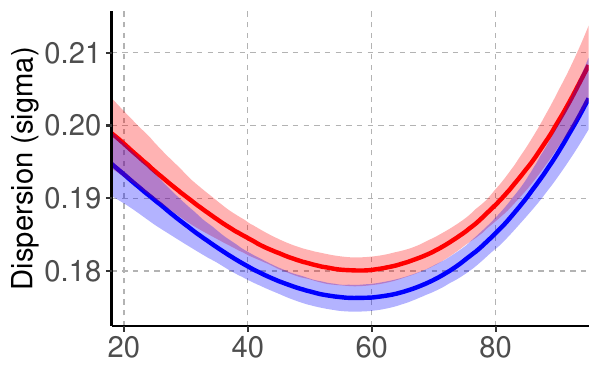}\\[1ex]
\includegraphics[width=\linewidth]{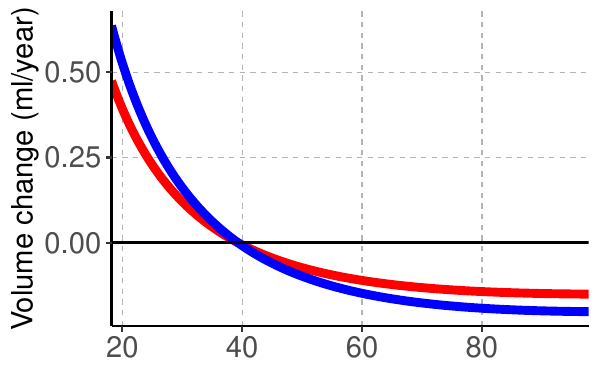}
\end{minipage}\hfill
\begin{minipage}{0.24\textwidth}
\centering
\footnotesize vertebrae L3\\[1ex]
\includegraphics[width=\linewidth]{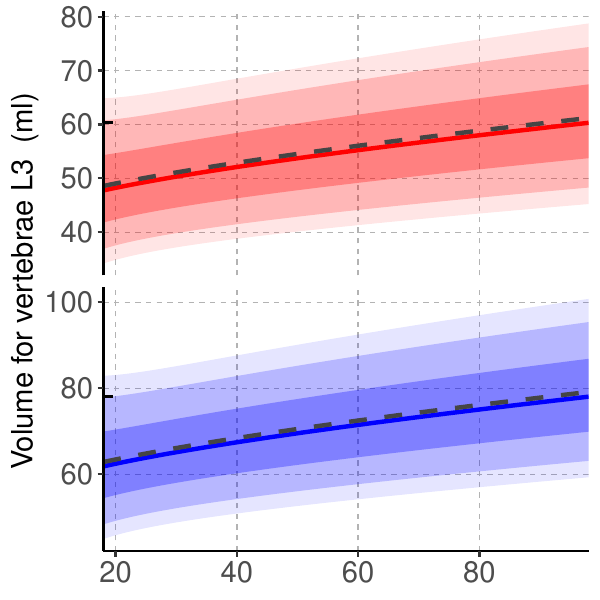}\\[1ex]
\includegraphics[width=\linewidth]{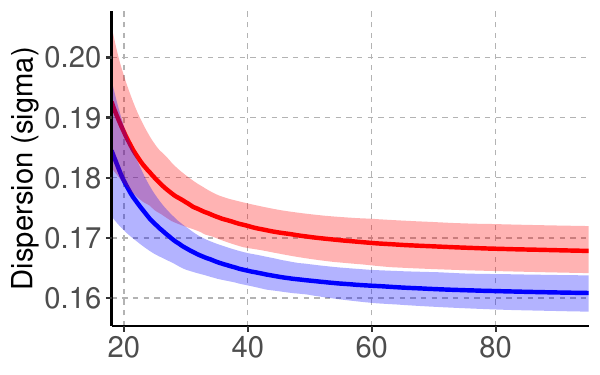}\\[1ex]
\includegraphics[width=\linewidth]{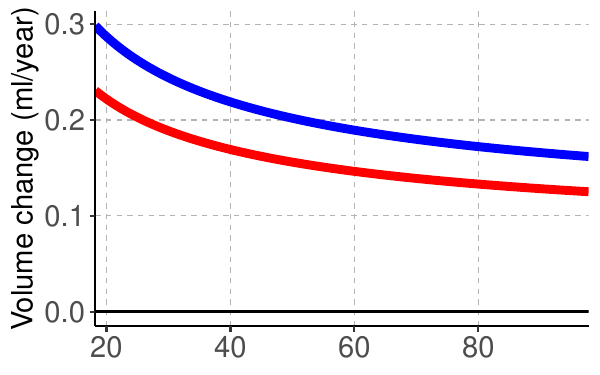}
\end{minipage}

\caption{(top) Reference body-volume trajectories estimated with GAMLSS, accounting for dataset-specific batch effects and stratified by sex (female, red; male, blue). Shaded ribbons show inter-centile ranges (5th--95th, 10th--90th, and 25th--75th). Dashed lines indicate the median for contrast-enhanced scans. Maximum predicted median volumes and their associated ages are marked on the curves. (middle) Absolute between-subject variability in volume. Sex-stratified bootstrap resampling of the generalized Gamma model; solid lines denote median bootstrap estimates and shaded areas the corresponding 95\% confidence intervals. (bottom) Rates of volumetric change (ml/year), computed as first derivatives of the sex-specific median GAMLSS trajectories. The horizontal line at zero indicates no change. Plots for all anatomical structures are provided in Supplement~S4.}
\label{fig:vol_gamlss_panels}
\end{figure*}

\begin{figure*}[p]
\centering

\begin{minipage}{0.24\textwidth}
\centering
\footnotesize liver\\[1ex]
\includegraphics[width=\linewidth]{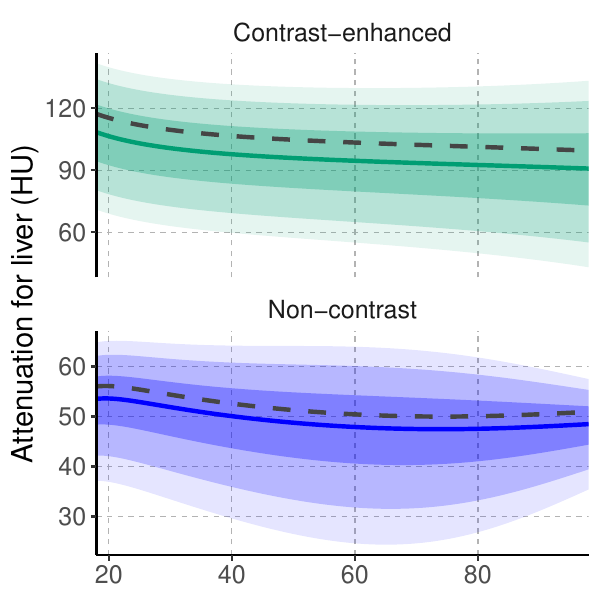}\\[1ex]
\includegraphics[width=\linewidth]{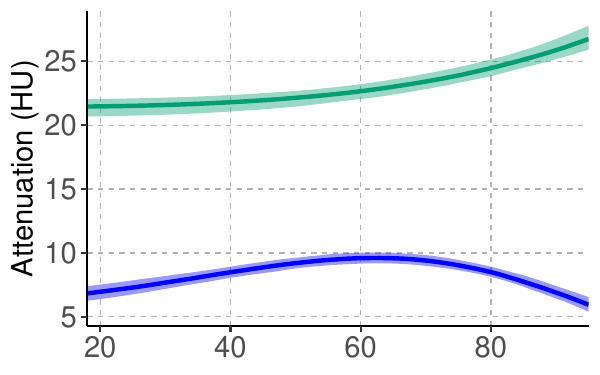}\\[1ex]
\includegraphics[width=\linewidth]{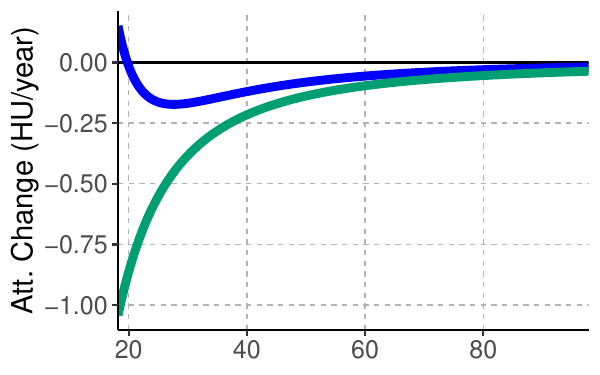}
\end{minipage}\hfill
\begin{minipage}{0.24\textwidth}
\centering
\footnotesize kidney right\\[1ex]
\includegraphics[width=\linewidth]{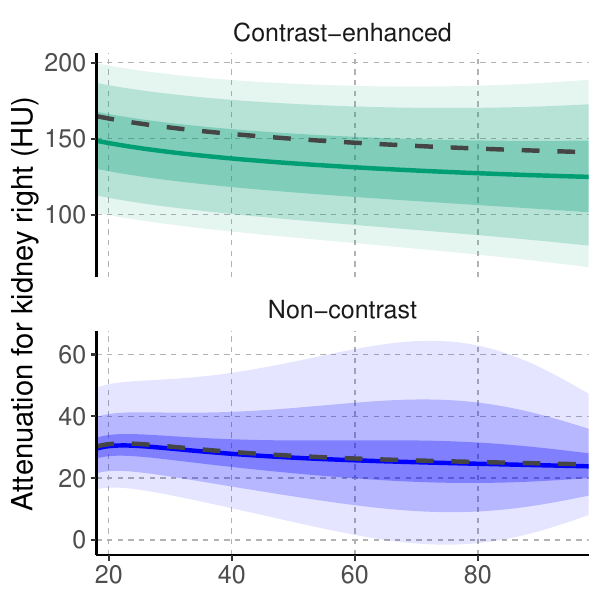}\\[1ex]
\includegraphics[width=\linewidth]{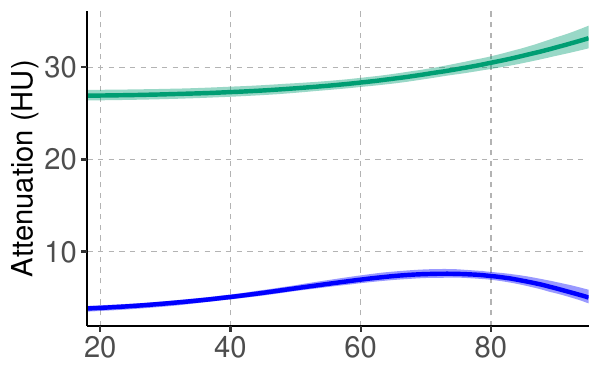}\\[1ex]
\includegraphics[width=\linewidth]{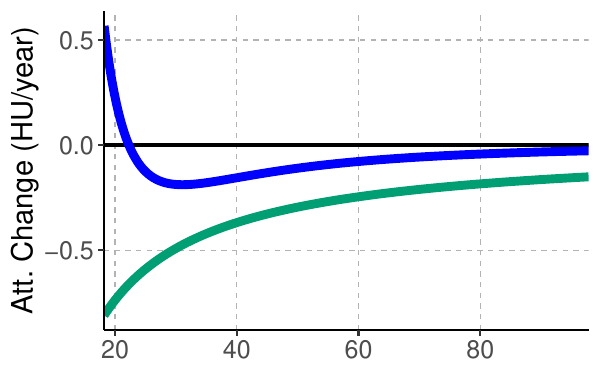}
\end{minipage}\hfill
\begin{minipage}{0.24\textwidth}
\centering
\footnotesize spleen\\[1ex]
\includegraphics[width=\linewidth]{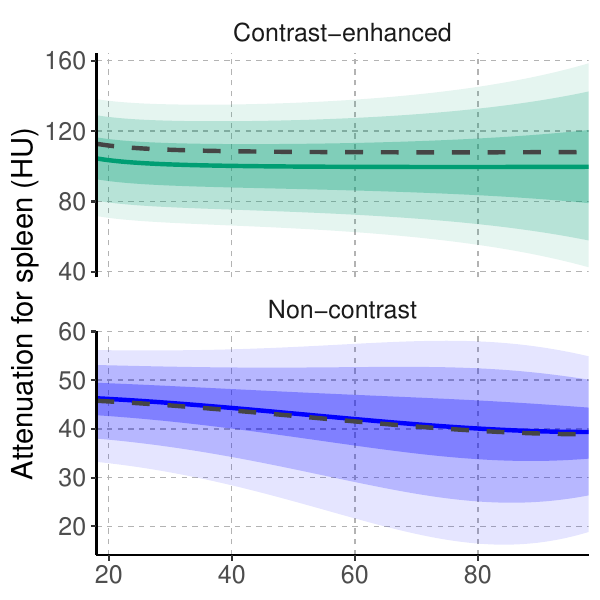}\\[1ex]
\includegraphics[width=\linewidth]{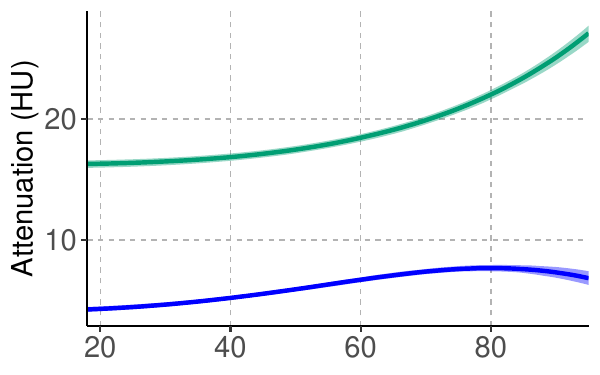}\\[1ex]
\includegraphics[width=\linewidth]{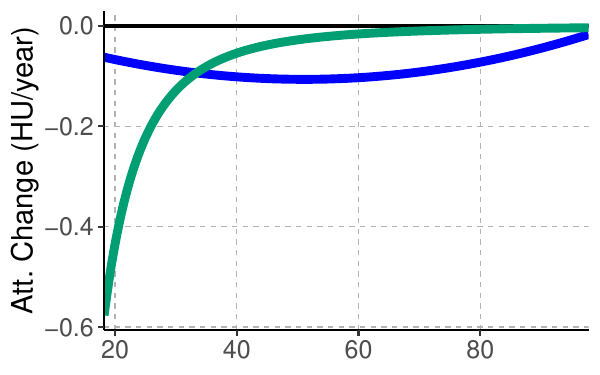}
\end{minipage}\hfill
\begin{minipage}{0.24\textwidth}
\centering
\footnotesize lung upper lobe right\\[1ex]
\includegraphics[width=\linewidth]{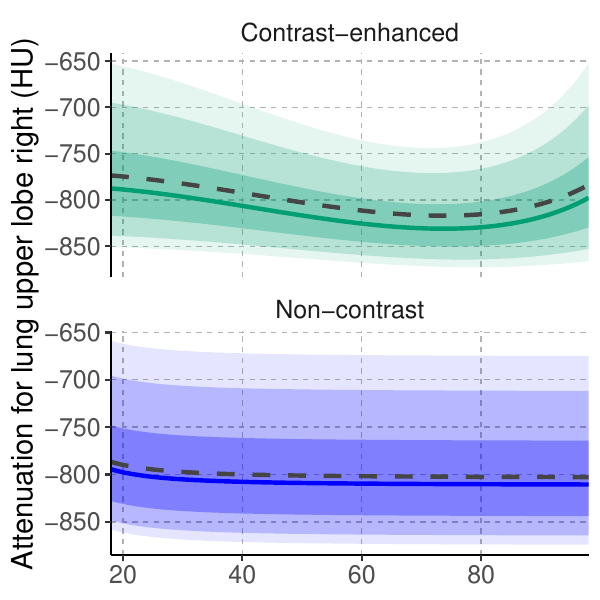}\\[1ex]
\includegraphics[width=\linewidth]{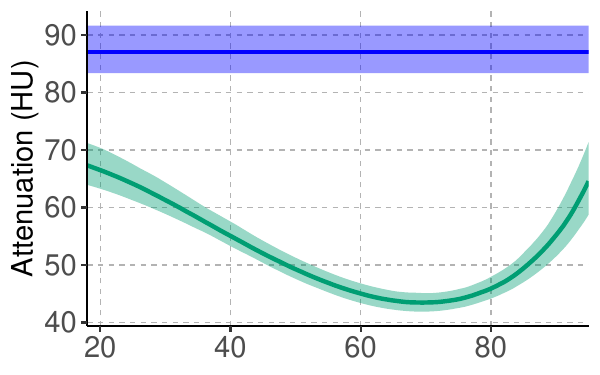}\\[1ex]
\includegraphics[width=\linewidth]{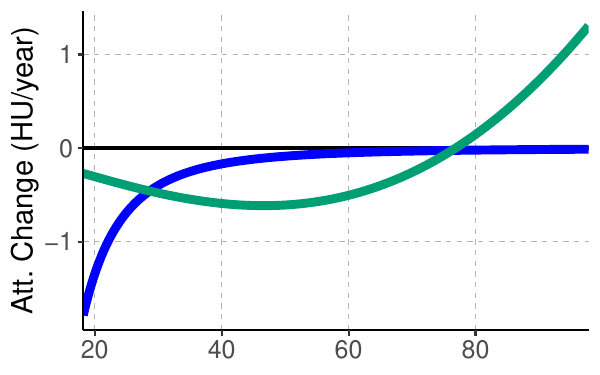}
\end{minipage}

\medskip

\begin{minipage}{0.24\textwidth}
\centering
\footnotesize aorta\\[1ex]
\includegraphics[width=\linewidth]{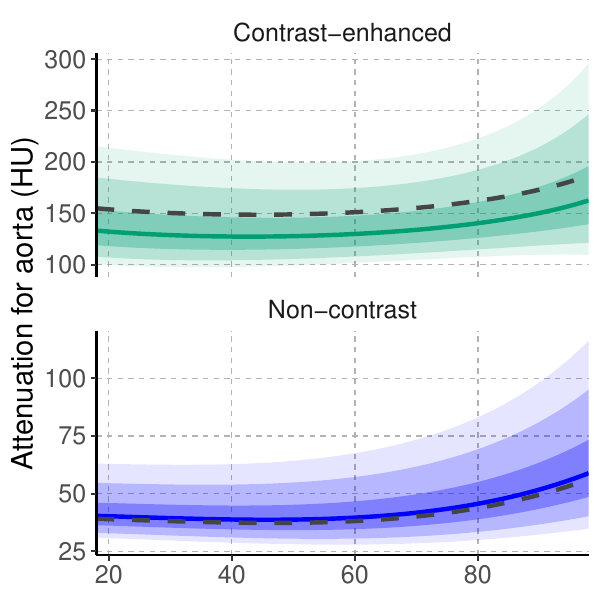}\\[1ex]
\includegraphics[width=\linewidth]{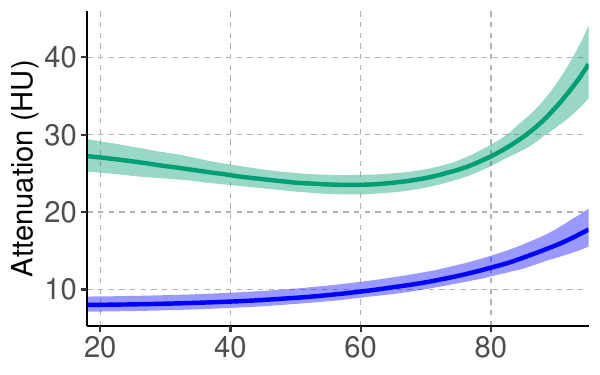}\\[1ex]
\includegraphics[width=\linewidth]{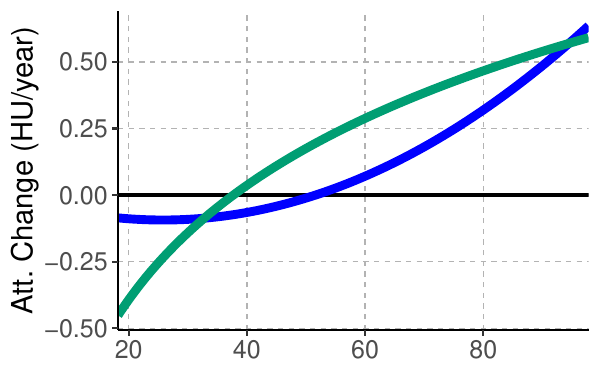}
\end{minipage}\hfill
\begin{minipage}{0.24\textwidth}
\centering
\footnotesize heart\\[1ex]
\includegraphics[width=\linewidth]{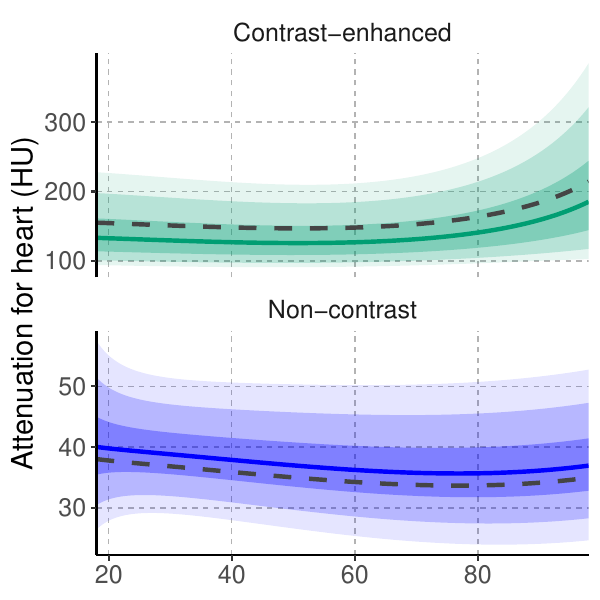}\\[1ex]
\includegraphics[width=\linewidth]{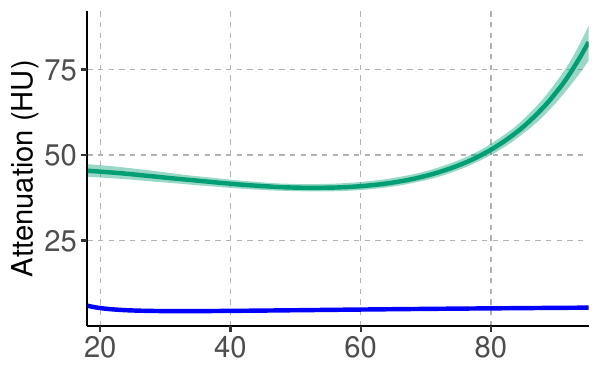}\\[1ex]
\includegraphics[width=\linewidth]{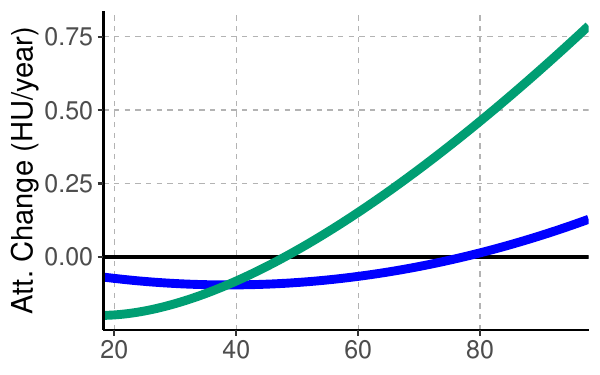}
\end{minipage}\hfill
\begin{minipage}{0.24\textwidth}
\centering
\footnotesize gluteus minimus right\\[1ex]
\includegraphics[width=\linewidth]{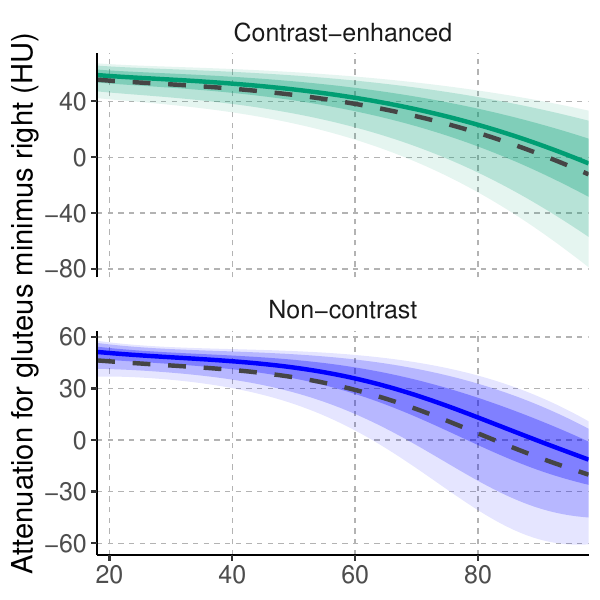}\\[1ex]
\includegraphics[width=\linewidth]{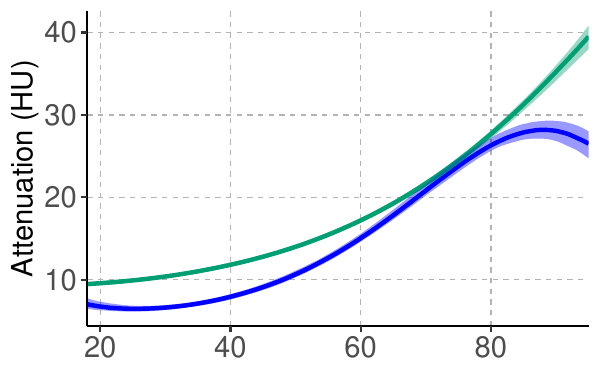}\\[1ex]
\includegraphics[width=\linewidth]{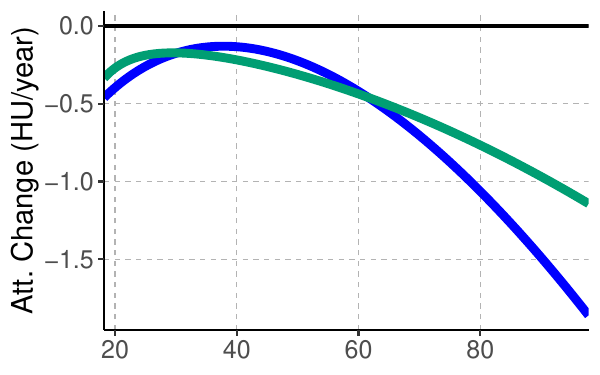}
\end{minipage}\hfill
\begin{minipage}{0.24\textwidth}
\centering
\footnotesize vertebrae L3\\[1ex]
\includegraphics[width=\linewidth]{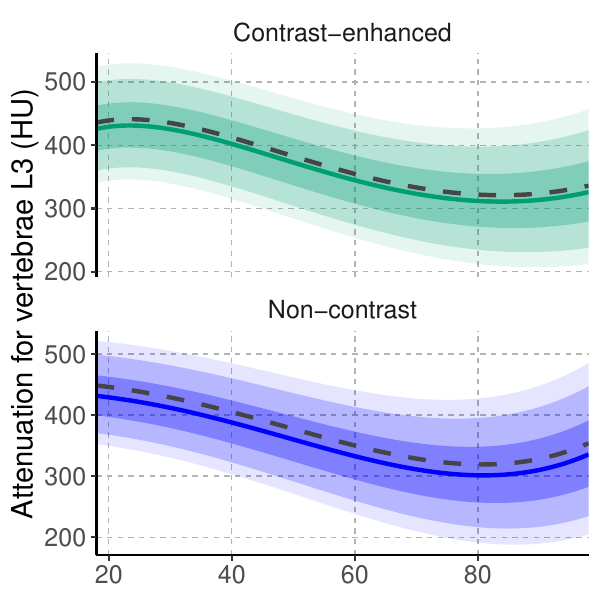}\\[1ex]
\includegraphics[width=\linewidth]{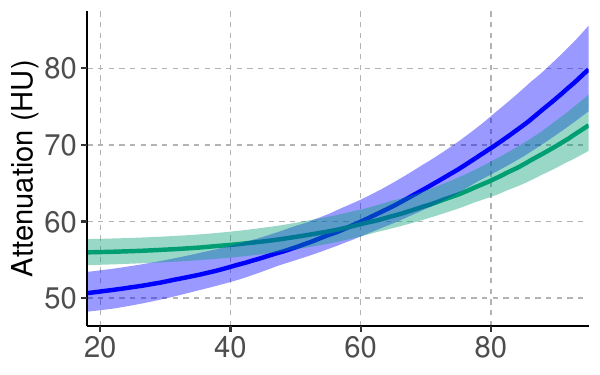}\\[1ex]
\includegraphics[width=\linewidth]{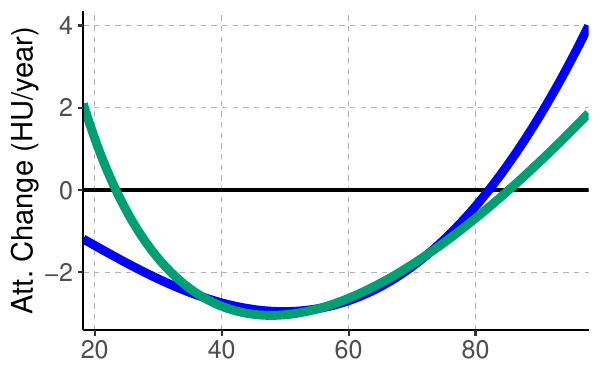}
\end{minipage}

\caption{(top) Reference attenuation trajectories estimated with GAMLSS and stratified by contrast. Solid lines indicate the median for men and dashed lines the median for women. Shaded ribbons show inter-centile ranges. (middle) Between-subject scale in attenuation estimated from GAMLSS. Lines show the median scale across 1{,}000 bootstrap resamples and shaded areas indicate the 95\% confidence interval. (bottom) Rates of attenuation change, computed as first derivatives of the median attenuation trajectories; the horizontal line at zero indicates no change. Plots for all anatomical structures are provided in Supplement~S5.}
\label{fig:int_gamlss_panels}
\end{figure*}

\subsection{Cross-sectional GAMLSS models}
Using the pathology-filtered cohort, we derived cross-sectional reference charts for both organ volume and CT attenuation across adulthood. 
We then fitted GAMLSS models to estimate full age-conditional reference distributions, using fractional polynomials to capture non-linear age effects (Methods Sec.~\ref{sec:methodCrossSec}). Volumes were modeled with a generalized Gamma (GG) family, whereas attenuation was modeled with the ST1 family to accommodate skewness and heavier tails in HU; attenuation models were estimated separately for contrast-enhanced and non-contrast examinations.
Figs.~\ref{fig:vol_gamlss_panels} and~\ref{fig:int_gamlss_panels} summarize the resulting cross-sectional reference charts for eight exemplar structures (full results in Supplement~S4--S5). Across structures, the estimated centiles show pronounced non-linear age trajectories. Fractional-polynomial selection frequently identified non-linear age terms for both location and, where required, scale (Table~\ref{tab:FP_all}), motivating distributional modeling beyond linear or Gaussian assumptions.

For volume, organ-specific aging trajectories differ qualitatively across tissues, with strong non-linear changes in the median and, for many structures, age-dependent dispersion. The age effect was significant after Bonferroni correction for all eight structures (Supplement~S3). 
Median volumes were consistently larger in men, with multiplicative factors ranging from 1.18 (liver) to 1.35 (lung upper lobe right), corresponding to approximately 18--35\% higher typical volumes (male vs.\ female). Between-subject variability also changed across adulthood for several organs, whereas for others dispersion was adequately captured without an additional age term (Table~\ref{tab:FP_all}).

For attenuation, the reference models of the eight exemplar structures (16 structure--contrast fits) showed four consistent patterns. First, attenuation varied non-linearly with age across organs and contrast states, reflected in the frequent selection of non-linear fractional-polynomial age terms for both location and, when required, scale (Table~\ref{tab:FP_all}). Second, sex differences were systematic: mean attenuation was higher in women across all exemplar fits. Third, acquisition factors had strong effects: attenuation decreased with higher tube potential in nearly all fits, and contrast status induced pronounced shifts in the reference distributions (Fig.~\ref{fig:int_gamlss_panels}). Finally, dispersion was not constant: several organs required age- and/or sex-dependent scale terms, indicating systematic changes in between-subject variability across adulthood (Supplement S3, S8, S9, Table~\ref{tab:FP_all}).

\begin{table}[t]
\footnotesize
\caption{Fractional polynomial (FP) specifications for non-linear age effects in the location ($\mu$) and scale ($\sigma$) components of the cross-sectional models. FP powers were selected from $\{-2,-1,-0.5,0,0.5,1,2,3\}$ by BIC. Volume models use the GG family. Attenuation models use the ST1 family and are stratified by contrast status ($\times$ non-contrast, $\checkmark$ contrast-enhanced). Here $x$ denotes age in years, coefficients removed for clarity. FPs for all structures are listed in Supplement S10--11.}
\label{tab:FP_all}
\centering
\begin{tabular}{p{3.6cm}cll}
\toprule
\textbf{Volume (GG)} &  & FP$_\mu$(age) & FP$_\sigma$(age) \\
\midrule
aorta &  & $x^{2} + x^{3}$ & $x^{2} + x^{3}$ \\
gluteus minimus right &  & $\log(x) + x^{0.5}$ & $x^{2} + x^{2}\,\log(x)$ \\
heart &  & $x^{-1} + x^{3}$ & $x^{3}$ \\
kidney right &  & $x^{-0.5} + x^{3}$ &  \\
liver &  & $x^{0.5} + x^{2}$ & $x^{3}$ \\
lung upper lobe right &  & $x^{2} + x^{3}$ & $x + x^{3}$ \\
spleen &  & $x^{-2} + x^{3}$ & $x^{2} + x^{3}$ \\
vertebrae L3 &  & $x^{0.5}$ & $x^{-2}$ \\
\midrule
\textbf{Attenuation (ST1)} & Contrast & FP$_\mu$(age) & FP$_\sigma$(age) \\
\midrule
aorta & $\times$ & $x^{2} + x^{3}$ & $x^{3}$ \\
aorta & $\checkmark$ & $x + x\,\log(x)$ & $x^{3} + x^{3}\,\log(x)$ \\
gluteus minimus r & $\times$ & $x + x^{2} + x^{2}\,\log(x)$ & $x^{-0.5} + x^{3} + x^{3}\,\log(x)$ \\
gluteus minimus r & $\checkmark$ & $x^{-2} + x^{3}$ & $x^{3} + x^{3}\,\log(x)$ \\
heart & $\times$ & $x^{2} + x^{3}$ & $x^{-2} + x^{-1}$ \\
heart & $\checkmark$ & $x^{2} + x^{2}\,\log(x)$ & $x^{3} + x^{3}\,\log(x)$ \\
kidney r & $\times$ & $x^{-2} + x^{-2}\,\log(x)$ & $x^{3} + x^{3}\,\log(x)$ \\
kidney r & $\checkmark$ & $\log(x)$ & $x^{3}$ \\
liver & $\times$ & $x^{-2} + x^{-2}\,\log(x)$ & $x^{2} + x^{3}$ \\
liver & $\checkmark$ & $x^{-1}$ & $x^{3}$ \\
lung upper lobe r & $\times$ & $x^{-2}$ &  \\
lung upper lobe r & $\checkmark$ & $x^{3} + x^{3}\,\log(x)$ & $x^{3} + x^{3}\,\log(x)$ \\
spleen & $\times$ & $x^{2} + x^{3}$ & $x^{3} + x^{3}\,\log(x)$ \\
spleen & $\checkmark$ & $x^{-2}$ & $x^{3}$ \\
vertebrae L3 & $\times$ & $x^{3} + x^{3}\,\log(x)$ & $x^{2}$ \\
vertebrae L3 & $\checkmark$ & $x + x\,\log(x) + x^{2}$ & $x^{3}$ \\
\bottomrule
\end{tabular}
\end{table}

\begin{figure*}[t]
  \centering

  \begin{subfigure}[t]{\textwidth}
    \centering
    \begin{subfigure}[t]{0.32\textwidth}
      \centering
      \includegraphics[width=\linewidth]{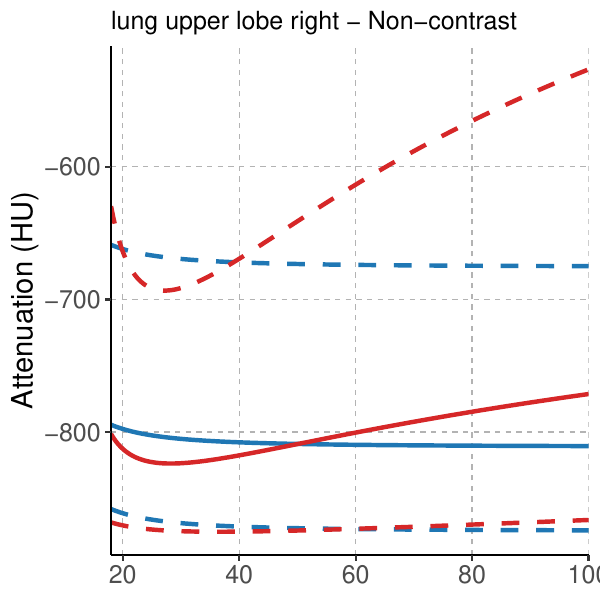}
    \end{subfigure}\hfill
    \begin{subfigure}[t]{0.32\textwidth}
      \centering
      \includegraphics[width=\linewidth]{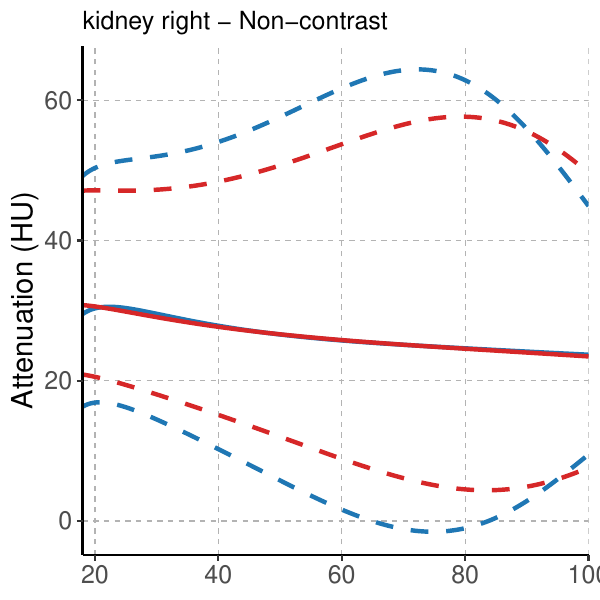}
    \end{subfigure}\hfill
    \begin{subfigure}[t]{0.32\textwidth}
      \centering
      \includegraphics[width=\linewidth]{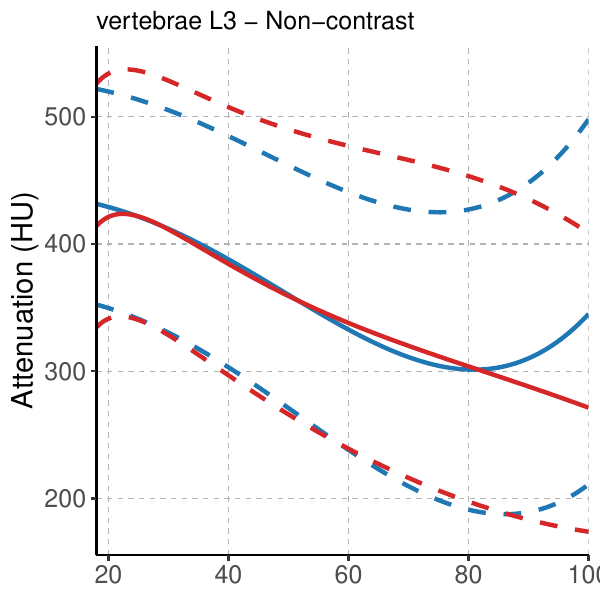}
    \end{subfigure}

    \caption{\textbf{Impact of LLM-filtering on Charts:} Filtered (blue) vs.\ non-filtered (red) ST1-GAMLSS centile trajectories (solid: p50; dashed: p5/p95).}
    \label{fig:impact_filtering_refcurves}
  \end{subfigure}

  \vspace{0.8em}

  \begin{subfigure}[t]{\textwidth}
    \centering
    \begin{subfigure}[t]{0.32\textwidth}
      \centering
      \includegraphics[width=\linewidth]{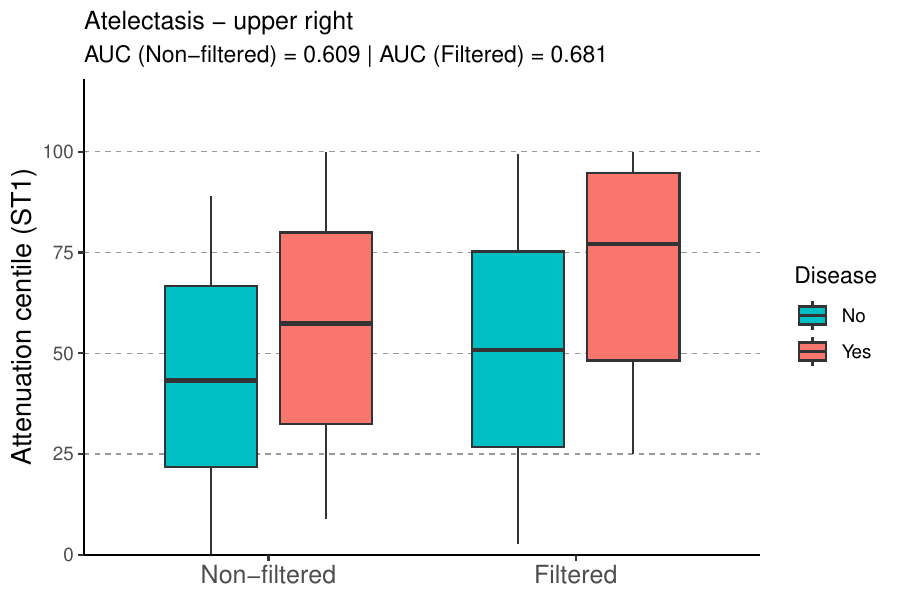}
    \end{subfigure}\hfill
    \begin{subfigure}[t]{0.32\textwidth}
      \centering
      \includegraphics[width=\linewidth]{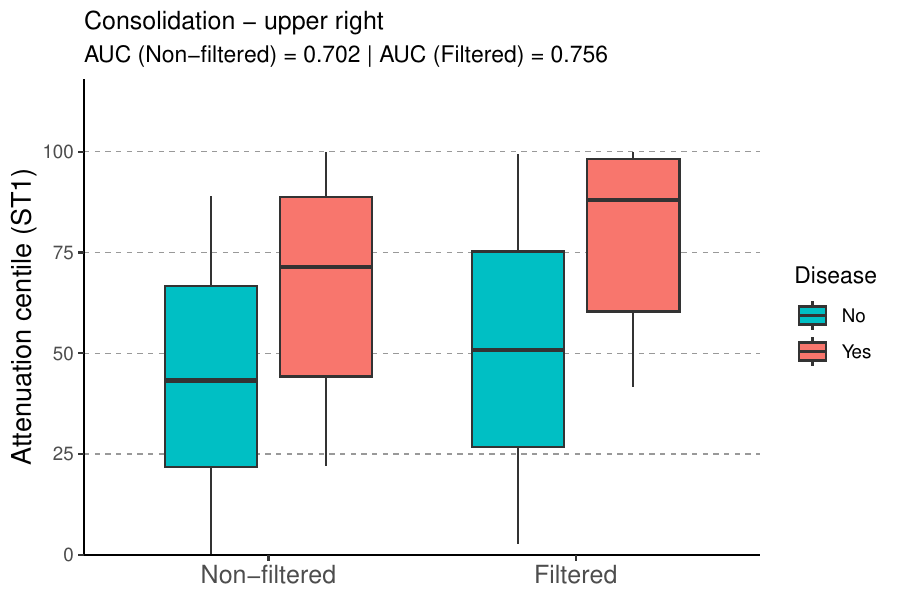}
    \end{subfigure}\hfill
    \begin{subfigure}[t]{0.32\textwidth}
      \centering
      \includegraphics[width=\linewidth]{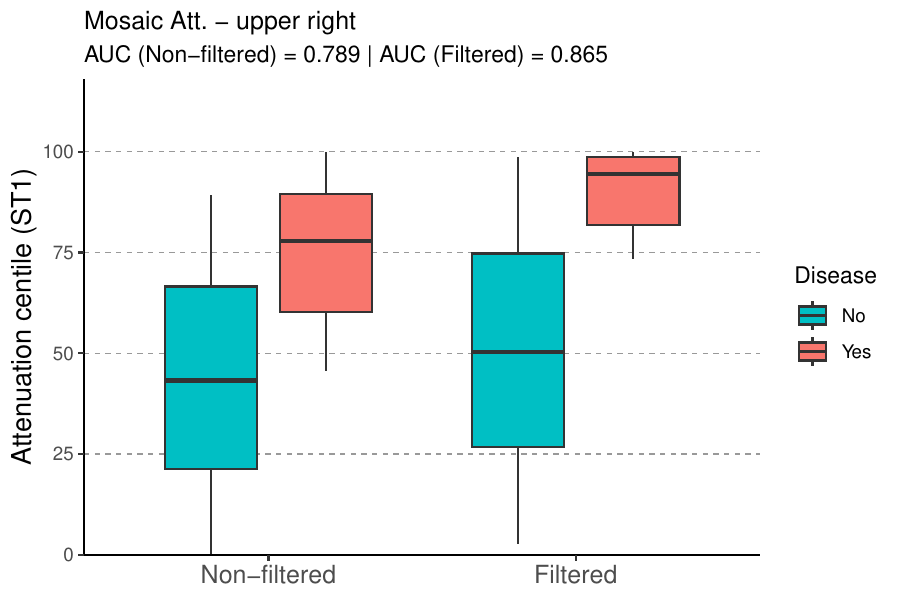}
    \end{subfigure}

    \caption{\textbf{Impact of LLM-filtering on Centile Scoring.} Boxplots compare non-filtered vs.\ filtered  centiles for the same subjects; subtitles report rank-based AUC.}
    \label{fig:impact_filtering_discrimination}
  \end{subfigure}

  \caption{\textbf{Report-based pathology filtering alters attenuation reference distributions and improves downstream centile discrimination.}
  \textbf{(A)} For three representative structures (male, non-contrast), filtered reference curves differ from non-filtered curves, with the largest differences in the distribution tails (p5/p95), which determine extreme centile scores.
  \textbf{(B)} In CT-Rate, applying filtered versus non-filtered models to the same subjects increases case--control separability for right upper lobe attenuation across three radiologic findings, indicating improved discriminability of individualized centile scores when the reference cohort is pathology-reduced.}
  \label{fig:impact_filtering_main}
\end{figure*}

Bootstrap resampling (1{,}000 iterations) indicated that estimated dispersion trajectories are stable across adulthood, with narrow pointwise 95\% intervals for the scale parameter $\sigma$ in both volume and attenuation models (Figs.~\ref{fig:vol_gamlss_panels}--\ref{fig:int_gamlss_panels}). 
{These intervals also make uneven age support transparent: age ranges with fewer observations, particularly younger adulthood in routine CT cohorts, can show wider uncertainty, especially for the dispersion parameter \(\sigma\), which influences the outer centiles.}
The derivative panels highlight ages at which organ-specific trajectories change direction or accelerate, providing an interpretable summary of non-linear aging dynamics.

{We further assessed whether the pooled reference trajectories were driven by individual source studies by analyzing the study-specific random effects in the GAMLSS models. We summarized these effects using best linear unbiased predictions (BLUPs), because they directly quantify the magnitude and direction of study-specific deviations from the pooled reference trajectory after accounting for the modeled covariates. For volume, BLUPs of the log-linked location parameter $\mu$ were expressed as percentage deviations from the pooled reference and were modest for most structures (median absolute deviation = 2.84\%, IQR 1.22--4.83\%). When examined by source study, median absolute deviations in $\mu$ were 4.43\% for Merlin, 3.89\% for CT-Rate, 3.18\% for INSPECT, 2.01\% for TotalSegmentator, and 1.63\% for TUM. Study effects were slightly larger for the log-linked dispersion parameter $\sigma$, but remained small for most structures (median absolute BLUP deviation = 3.70\%, IQR 1.15--8.57\%). For attenuation, study effects were evaluated separately for non-contrast and contrast-enhanced models. Because the ST1 attenuation models used an identity link for $\mu$, study-specific deviations were expressed in HU. Median absolute deviations in $\mu$ were 3.51 HU (IQR 0.00--17.0 HU) for non-contrast models and 9.30 HU (IQR 1.58--20.7 HU) for contrast-enhanced models.  For the log-linked attenuation dispersion parameter $\sigma$, deviations were expressed as percentage changes in dispersion and were moderate for most structures, 
median absolute deviations of 4.08\% (IQR 0.005--10.2\%) for non-contrast models and 4.03\% (IQR 1.22--11.4\%) for contrast-enhanced models. Larger deviations occurred for selected structures sensitive to contrast timing, scan coverage, or physiological state, including the urinary bladder, vessels, lung lobes, ribs, and esophagus. Together, these analyses indicate that the pooled trajectories were generally robust across source studies after accounting for age, sex, and acquisition parameters, while also identifying selected structures for which study-specific effects were more pronounced.}

Fig.~\ref{fig:impact_filtering_refcurves} illustrates how report-based pathology filtering alters attenuation reference charts. 
Centile trajectories derived from the pathology-reduced cohort differ from those obtained without filtering, with changes evident in the median trajectory and, more prominently, in the lower and upper tail centiles (p5 and p95). These shifts indicate that residual pathology in unfiltered clinical cohorts can broaden and displace the distribution tails, which directly affects the interpretation of extreme centile scores.

Attenuation distributions were frequently non-Gaussian, motivating the ST1 family for reliable centile estimation (Supplement~S2.5). For example, lung attenuation is strongly right-skewed ($\nu=14.3$), whereas iliopsoas and esophagus are left-skewed, and several structures (lung, aorta, esophagus) show very heavy tails ($\tau<2$). These features would distort tail centiles under Gaussian assumptions, whereas ST1 explicitly models both skewness and tail-heaviness, yielding improved fits.

\subsection{Individualized centile scores}
\begin{figure}[!tbp]
  \centering
  \includegraphics[width=0.9\textwidth]{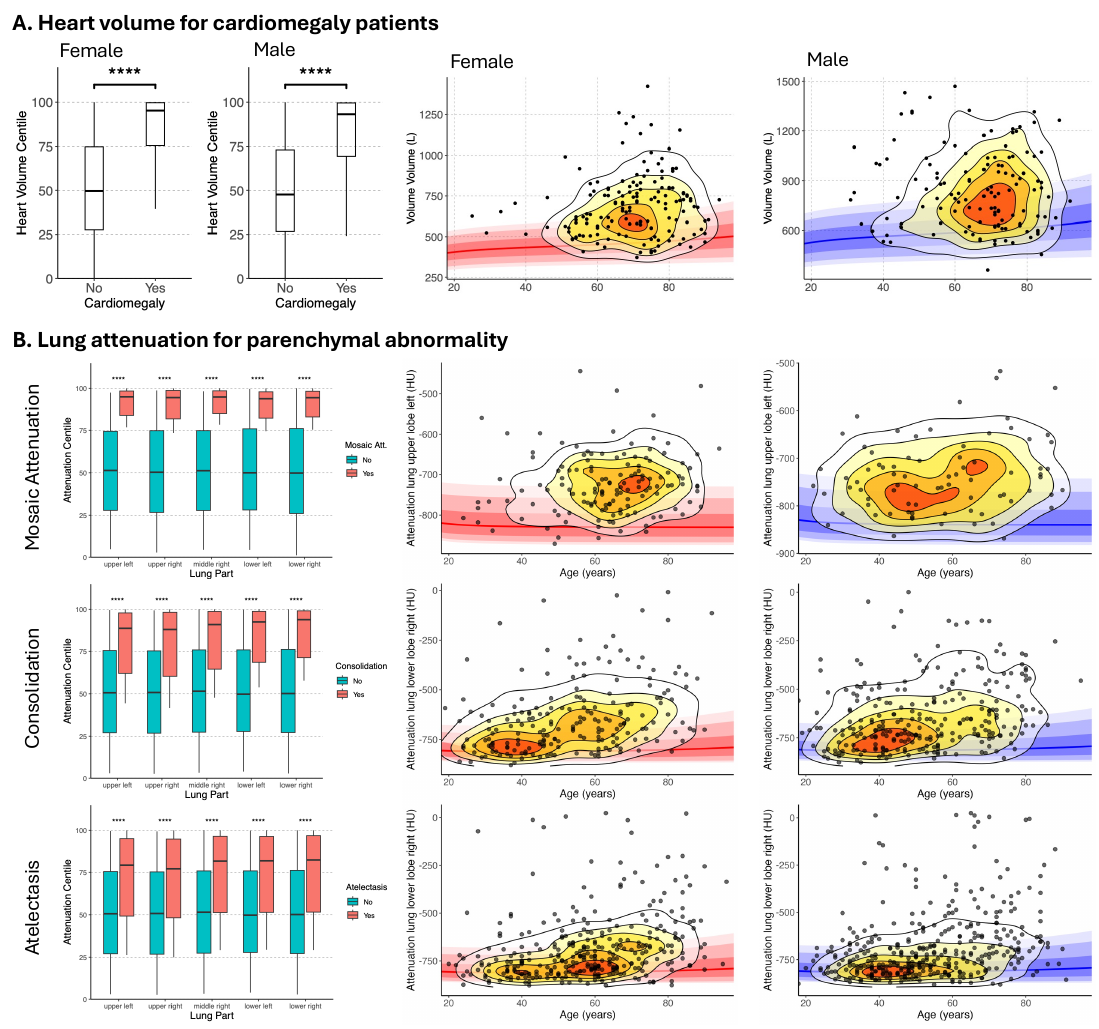}
\caption{\label{fig:DX}
Individualized centile score analysis for (A) heart volume in cardiomegaly patients, and (B) lung-parenchyma attenuation in participants with mosaic attenuation, consolidation, and atelectasis. For both panels, \textbf{left} box-and-whisker plots summarize the distribution of individualized centile scores stratified; between-group differences were assessed with a two-sided Wilcoxon rank-sum test, where \textasteriskcentered\textasteriskcentered\textasteriskcentered\textasteriskcentered\ indicates $P<0.0001$. \textbf{Right} panels show sex-specific reference centile curves (heart volume and lung lobe attenuation) with affected participants overlaid as a two-dimensional kernel-density estimate (yellow-to-red scale) and a randomly selected subset of affected individuals shown as black dots for visibility; the density estimate is computed using all patients.}
\end{figure}

Individualized centile scores place each organ-specific measurement into an age-, sex-, and acquisition-adjusted reference distribution, enabling direct interpretation of where an examination lies relative to expected variation. We illustrate this with two report-derived use cases from CT-Rate (Fig.~\ref{fig:DX}).

For organ volume, participants with cardiomegaly ($n=1{,}243$) were compared with an equal-sized control group without cardiomegaly. Heart-volume centile scores were significantly higher in the cardiomegaly group in both women and men (Fig.~\ref{fig:DX}A, left), and affected participants concentrated toward the upper tail of the sex-specific reference charts (Fig.~\ref{fig:DX}A, right), consistent with volumetric enlargement.
For organ attenuation, participants with lung parenchyma findings (mosaic attenuation, consolidation, or atelectasis) were compared with controls without these findings. Across all five lung lobes, cases showed significantly upward-shifted attenuation centile scores (Fig.~\ref{fig:DX}B, left) and more frequent occupancy of high-centile regions when overlaid on the reference charts (Fig.~\ref{fig:DX}B, right), indicating elevated attenuation relative to normative expectations.
Together, these examples demonstrate that centile scoring provides a unified, covariate-adjusted representation of both volumetric and attenuation phenotypes while preserving sensitivity to clinically relevant deviations. 

Moreover, this sensitivity depends on the reference cohort: applying filtered versus non-filtered attenuation references to the same subjects increased rank-based AUC for right upper lobe attenuation from 0.609 to 0.681 (atelectasis), from 0.702 to 0.756 (consolidation), and from 0.789 to 0.865 (mosaic attenuation) (Fig.~\ref{fig:impact_filtering_discrimination}). Because the underlying subjects are held fixed and only the  reference model changes, these gains quantify the downstream benefit of report-based pathology exclusion for centile scoring.

\begin{figure*}[th]
\centering

\begin{minipage}{0.24\textwidth}
\centering
\footnotesize liver\\[1ex]
\includegraphics[width=\linewidth]{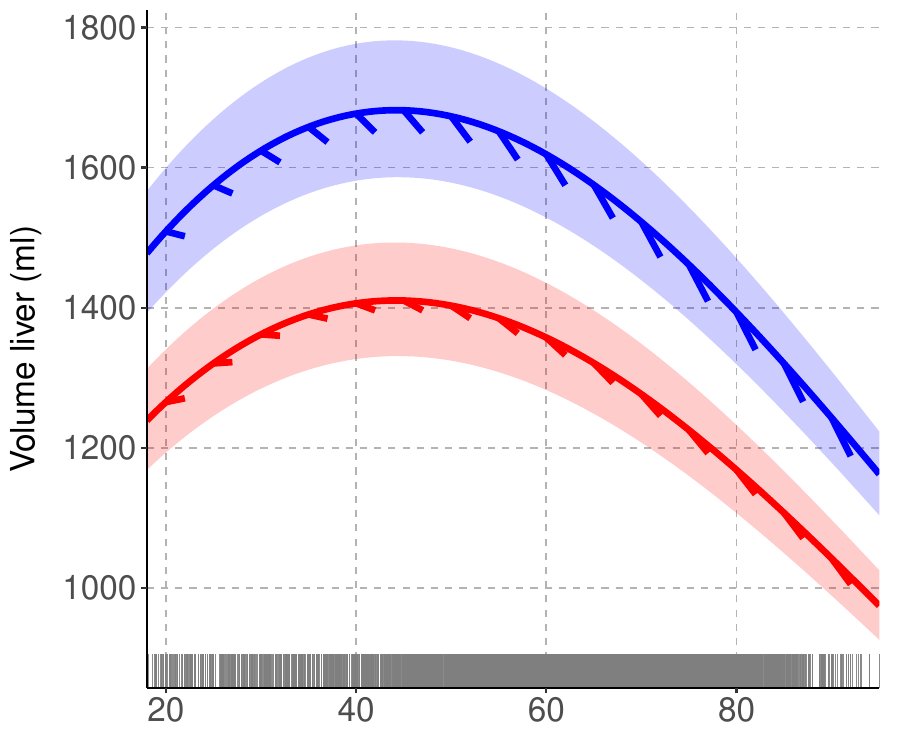}
\end{minipage}\hfill
\begin{minipage}{0.24\textwidth}
\centering
\footnotesize kidney right\\[1ex]
\includegraphics[width=\linewidth]{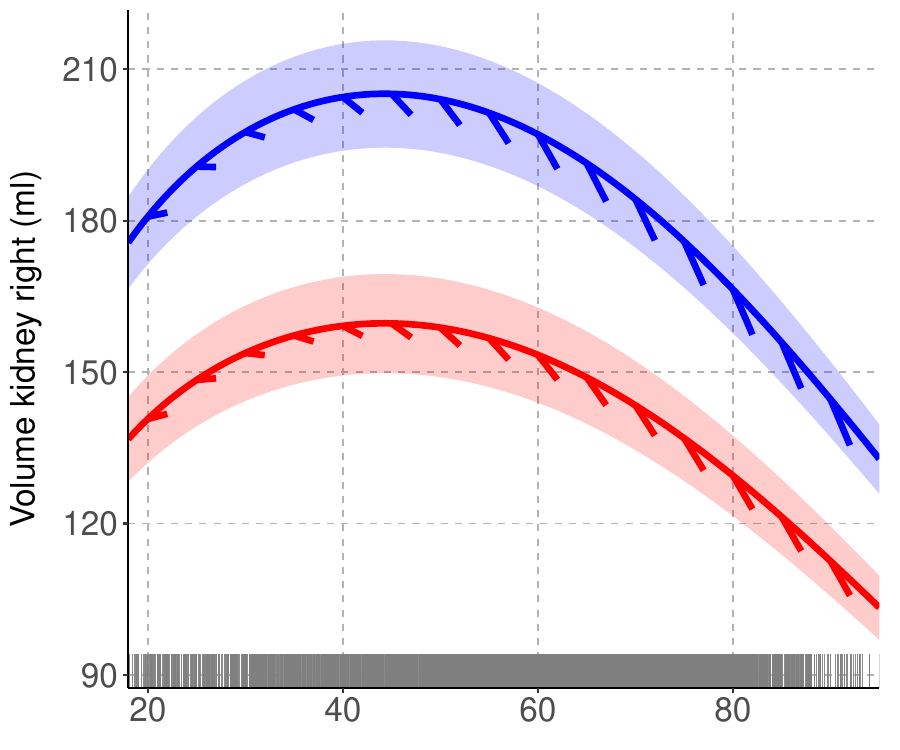}
\end{minipage}\hfill
\begin{minipage}{0.24\textwidth}
\centering
\footnotesize spleen\\[1ex]
\includegraphics[width=\linewidth]{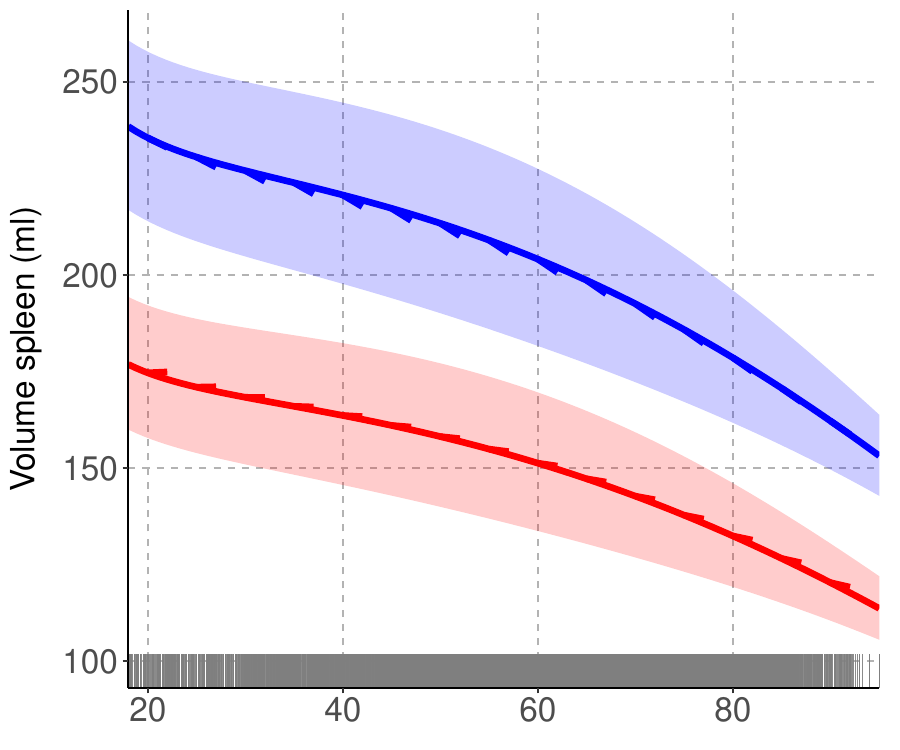}
\end{minipage}\hfill
\begin{minipage}{0.24\textwidth}
\centering
\footnotesize lung upper lobe right\\[1ex]
\includegraphics[width=\linewidth]{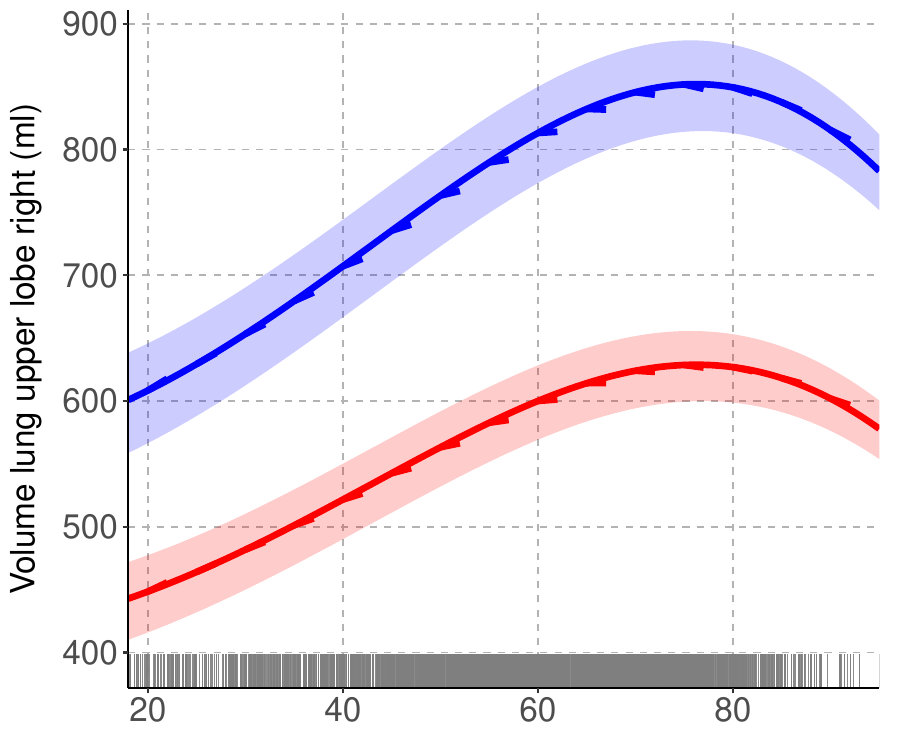}
\end{minipage}

\medskip

\begin{minipage}{0.24\textwidth}
\centering
\footnotesize aorta\\[1ex]
\includegraphics[width=\linewidth]{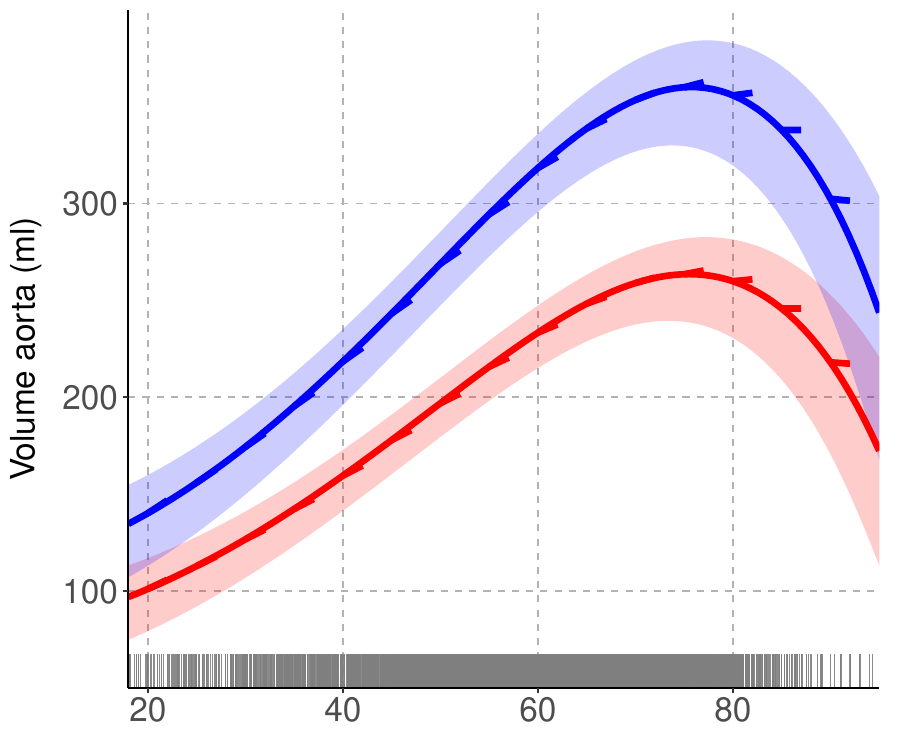}
\end{minipage}\hfill
\begin{minipage}{0.24\textwidth}
\centering
\footnotesize heart\\[1ex]
\includegraphics[width=\linewidth]{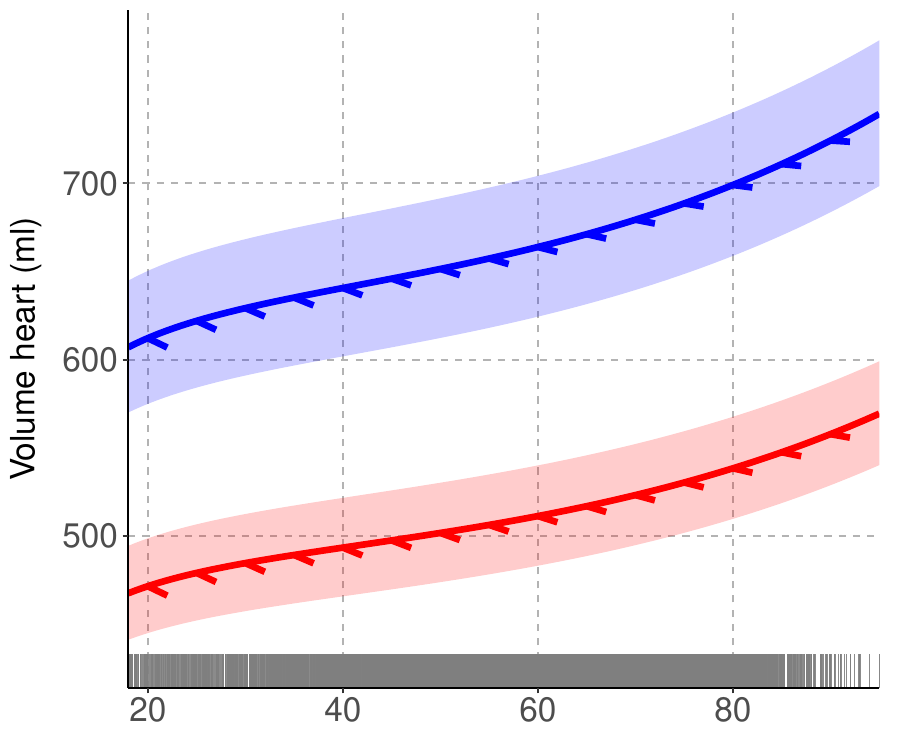}
\end{minipage}\hfill
\begin{minipage}{0.24\textwidth}
\centering
\footnotesize gluteus minimus right\\[1ex]
\includegraphics[width=\linewidth]{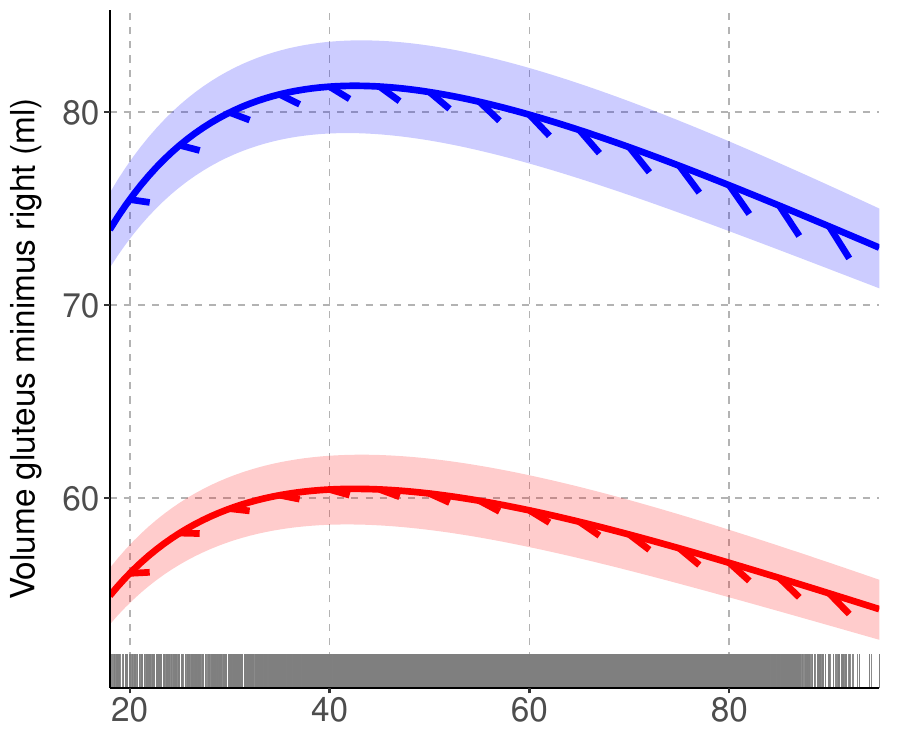}
\end{minipage}\hfill
\begin{minipage}{0.24\textwidth}
\centering
\footnotesize vertebrae L3\\[1ex]
\includegraphics[width=\linewidth]{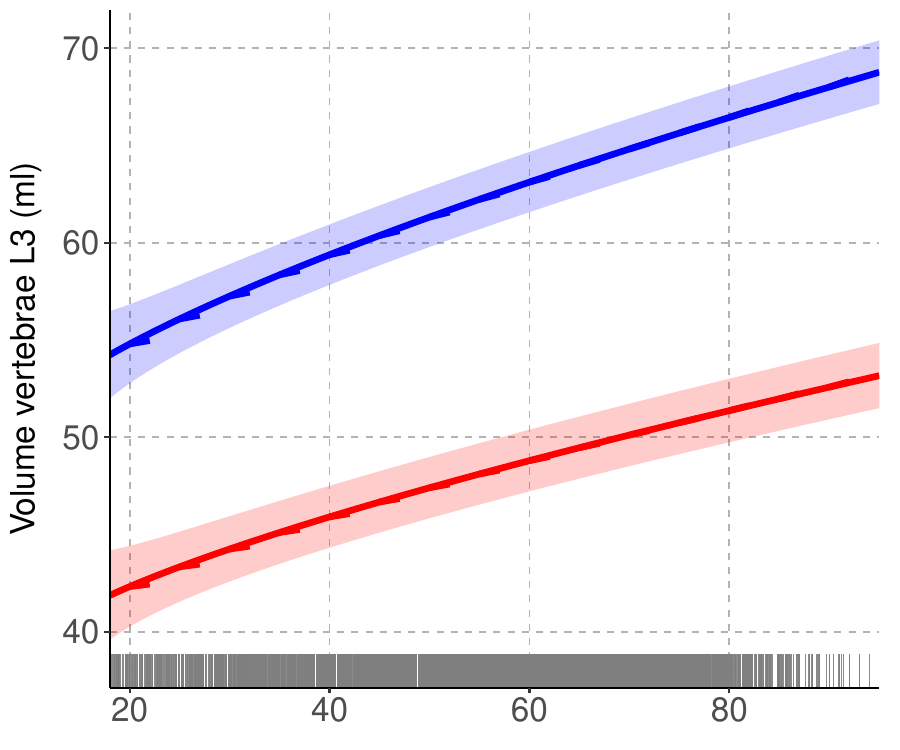}
\end{minipage}

\caption{Longitudinal volume GAMLSS fits by organ. Curves show baseline organ volume as a function of baseline age for women (red) and men (blue); shaded ribbons indicate the 25th-75th centile range. Line segments depict estimated within-subject change in volume over two years (time from baseline). Interactions between time from baseline, baseline age, and sex allow the slopes of the line segments to vary accordingly, combining cross-sectional age effects with short-term longitudinal volume change. Rug marks baseline ages of participants with follow-up (gray).
}
\label{fig:long_vol}
\end{figure*}

\begin{figure*}
\centering

\begin{minipage}{0.24\textwidth}
\centering
\footnotesize liver\\[1ex]
\includegraphics[width=\linewidth]{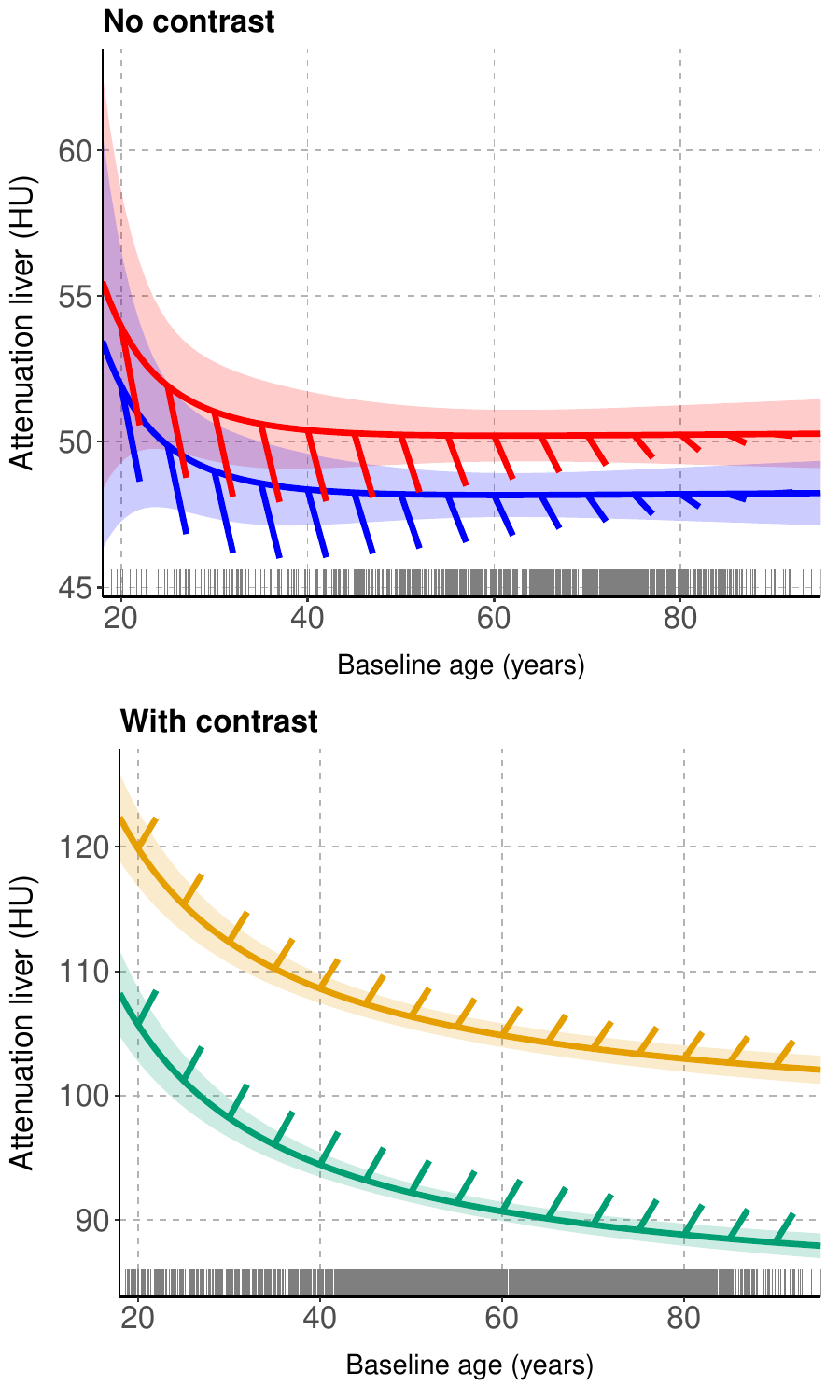}
\end{minipage}\hfill
\begin{minipage}{0.24\textwidth}
\centering
\footnotesize kidney right\\[1ex]
\includegraphics[width=\linewidth]{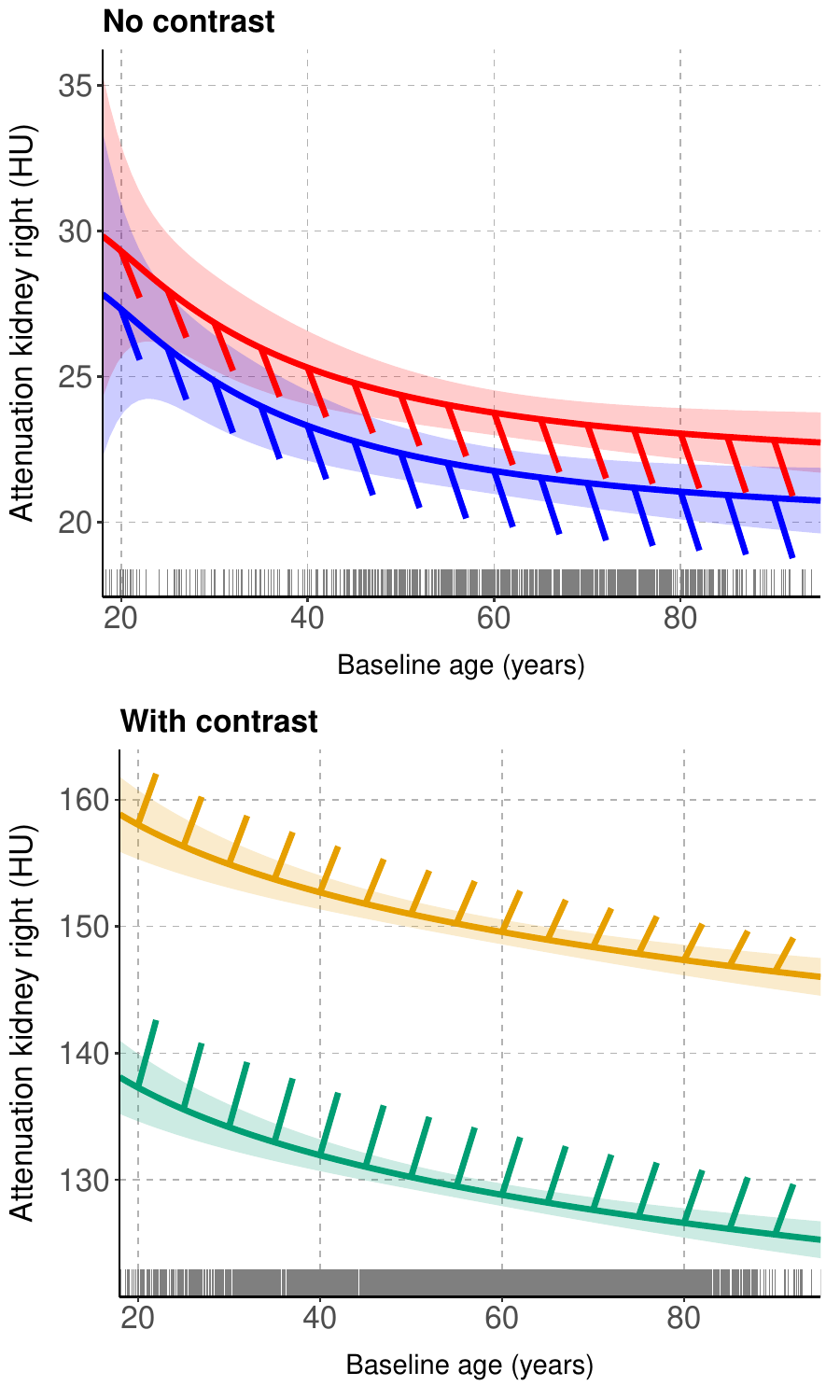}
\end{minipage}\hfill
\begin{minipage}{0.24\textwidth}
\centering
\footnotesize spleen\\[1ex]
\includegraphics[width=\linewidth]{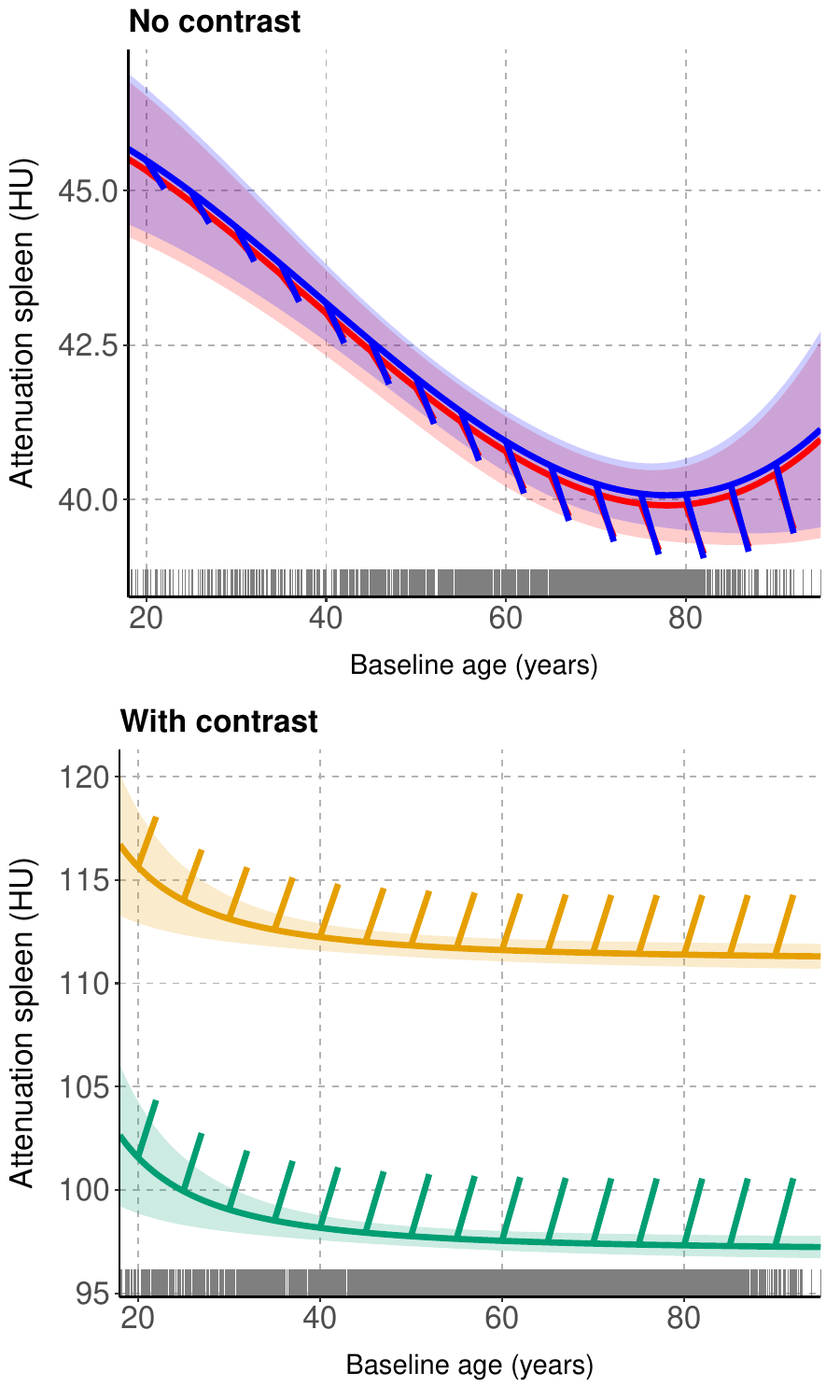}
\end{minipage}\hfill
\begin{minipage}{0.24\textwidth}
\centering
\footnotesize lung upper lobe right\\[1ex]
\includegraphics[width=\linewidth]{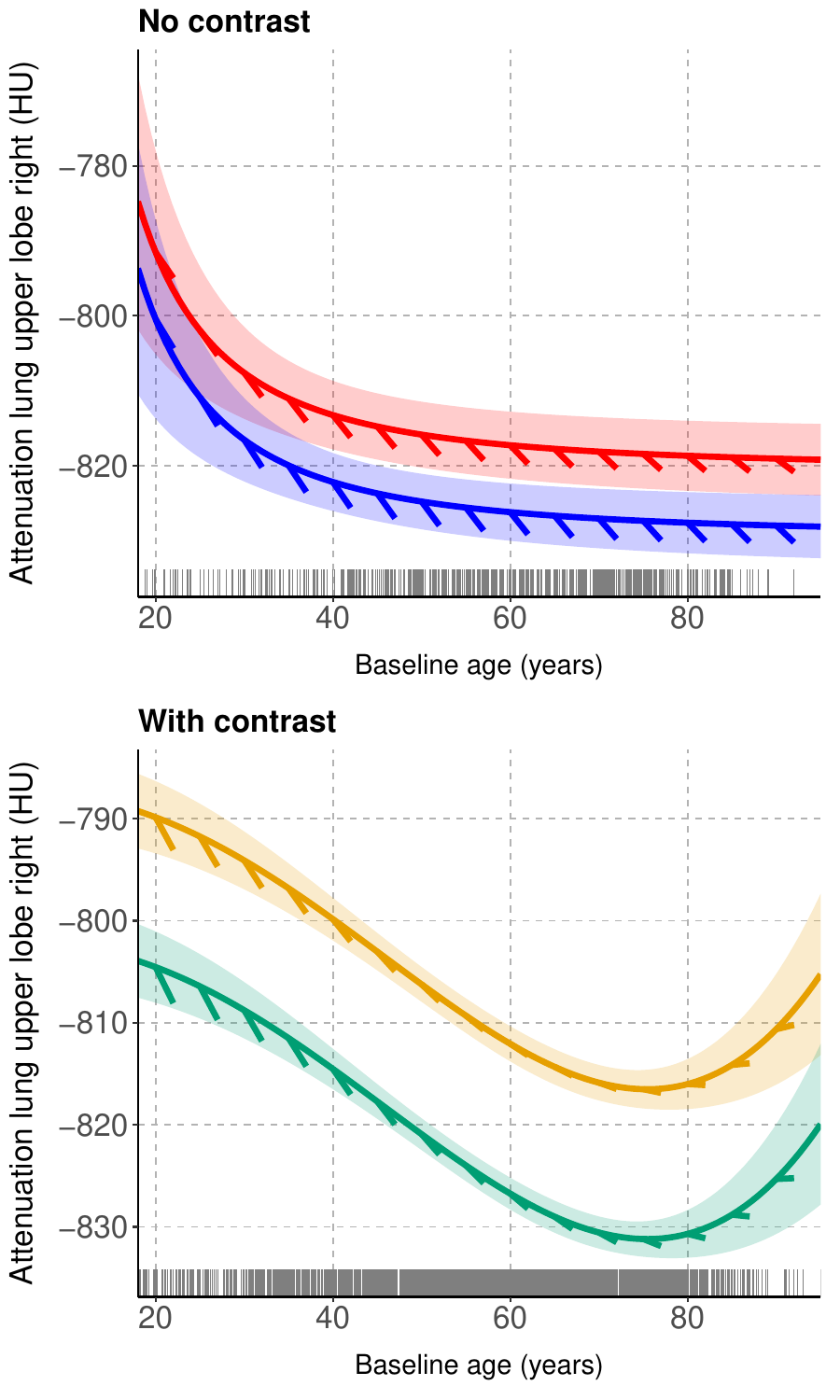}
\end{minipage}

\medskip

\begin{minipage}{0.24\textwidth}
\centering
\footnotesize aorta\\[1ex]
\includegraphics[width=\linewidth]{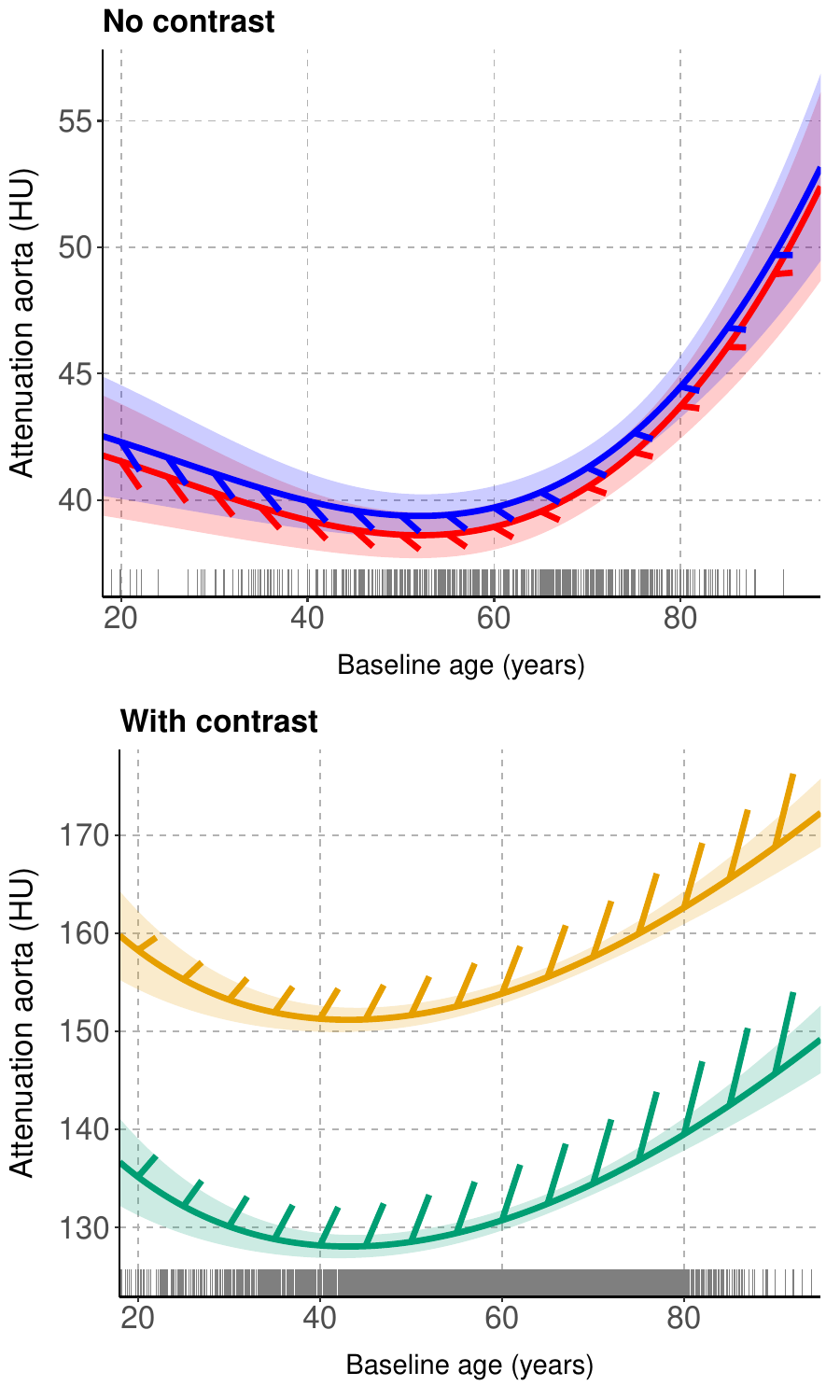}
\end{minipage}\hfill
\begin{minipage}{0.24\textwidth}
\centering
\footnotesize heart\\[1ex]
\includegraphics[width=\linewidth]{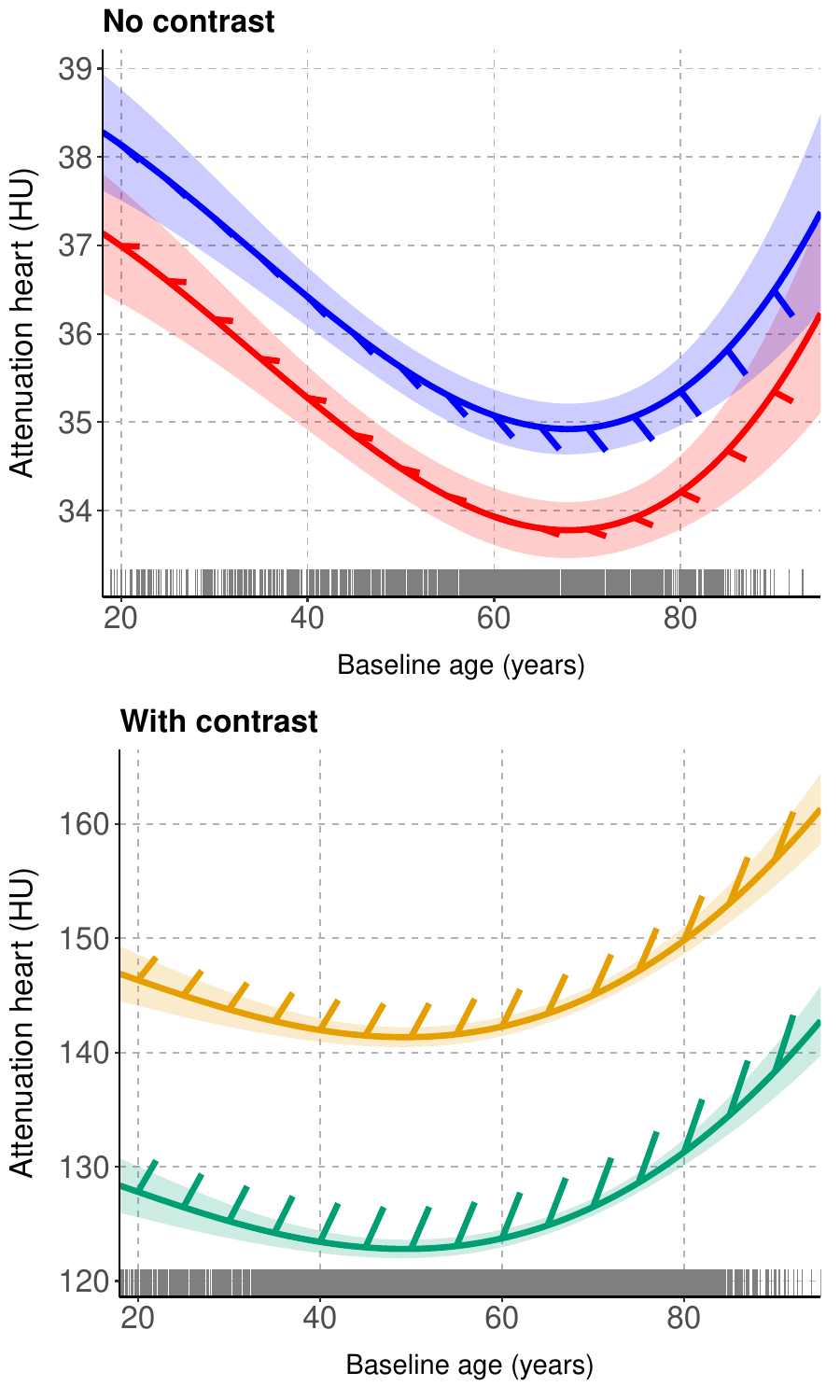}
\end{minipage}\hfill
\begin{minipage}{0.24\textwidth}
\centering
\footnotesize gluteus minimus right\\[1ex]
\includegraphics[width=\linewidth]{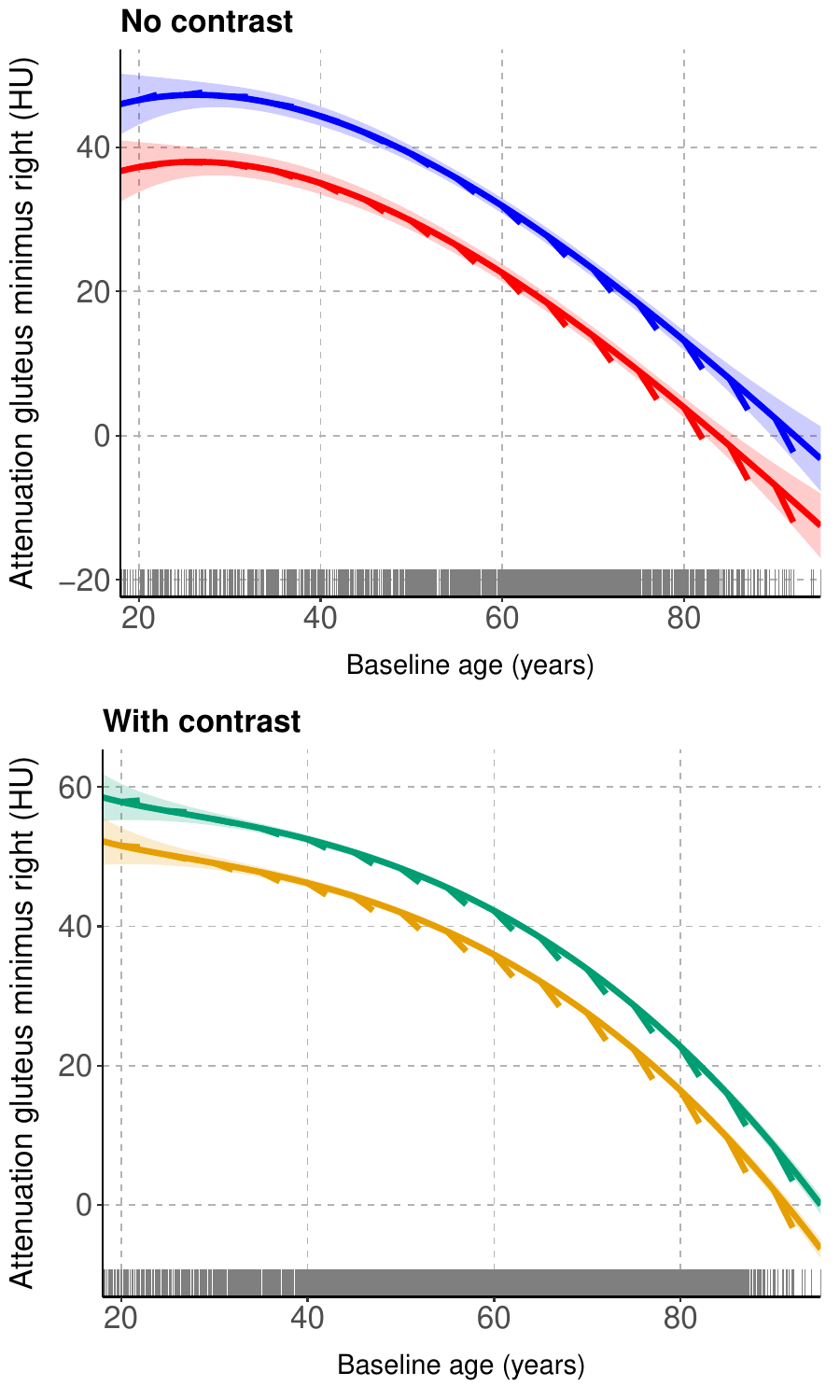}
\end{minipage}\hfill
\begin{minipage}{0.24\textwidth}
\centering
\footnotesize vertebrae L3\\[1ex]
\includegraphics[width=\linewidth]{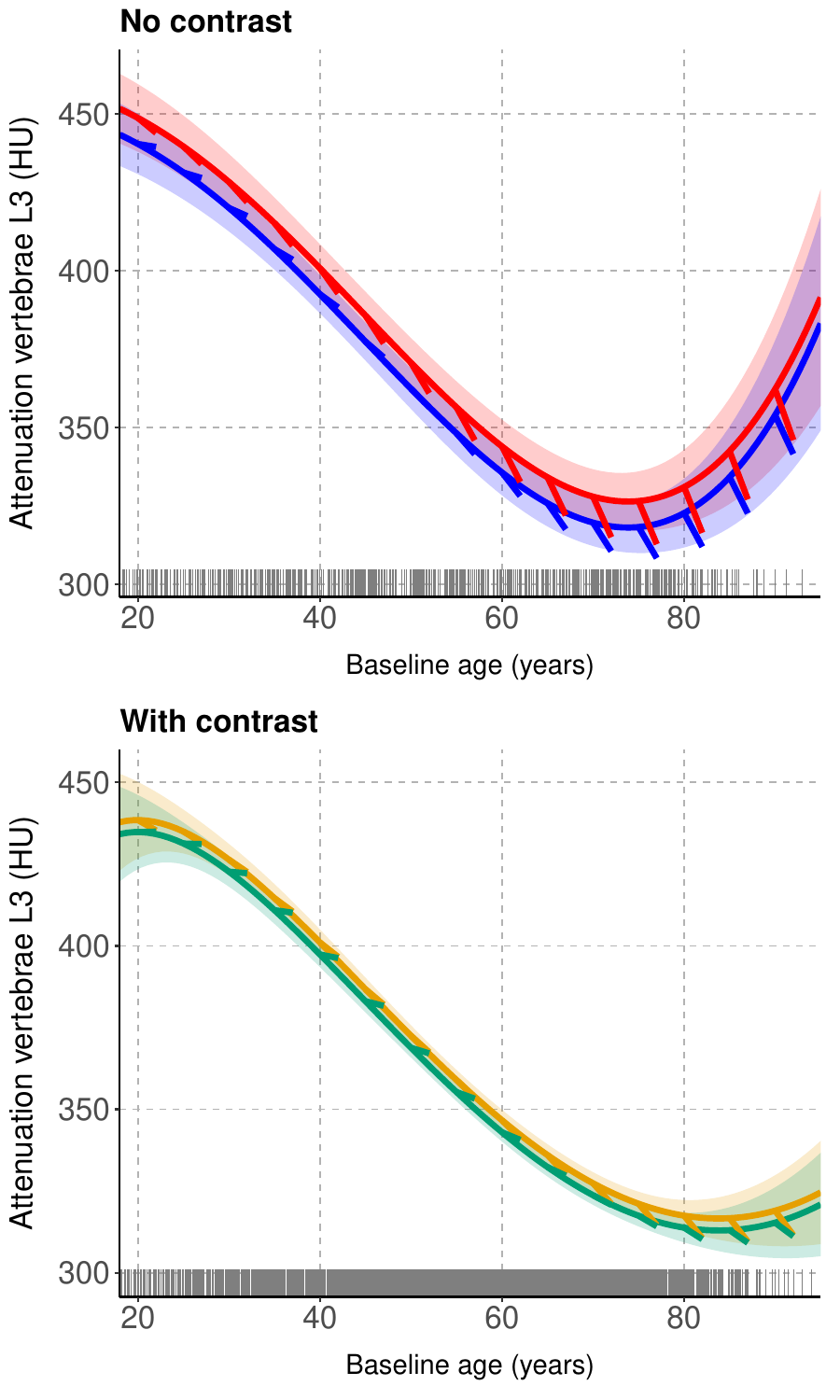}
\end{minipage}

\caption{Longitudinal attenuation GAMM fits for selected organs. 
Separate panels show non-contrast (top) and contrast-enhanced (bottom) scans.
Curves show baseline CT attenuation as a function of baseline age for men (non-contrast blue, contrast-enhanced green) and women (non-contrast red, contrast-enhanced yellow), with shaded ribbons indicating 95\% confidence intervals.
Line segments represent the estimated within-subject change in attenuation over two years. 
Interactions of time from baseline with baseline age and sex allow segment slopes to vary, integrating cross-sectional baseline-age effects with longitudinal attenuation change. 
Rug marks the baseline ages of participants with follow-up scans (gray).}
\label{fig:long_int}
\end{figure*}

\subsection{Longitudinal volume and attenuation models}
Longitudinal analyses were restricted to the TUM cohort with repeated examinations per participant. 
The age is decomposed as $\mathrm{age}=\mathrm{age}_b+\mathrm{time}_b$ (Methods \ref{sec:long}), with baseline age, $\mathrm{age}_b$, and time from baseline, $\mathrm{time}_b$.
Figs.~\ref{fig:long_vol} and \ref{fig:long_int}  compare the cross-sectional baseline trajectories with the model-implied longitudinal changes, depicted as line segments, to illustrate within-subject changes over time (all plots in Supplement~S6--S7).
Fixed-effect coefficients are reported in Table~\ref{tab:long_models} (full results in Supplement~S12--S13).

\paragraph{Longitudinal volume (GAMLSS--GG).}
Longitudinal organ volume was modeled with mixed-effects GAMLSS (GG family), adding $\mathrm{time}_b$ and its interactions with baseline age and sex to the cross-sectional specification. 
In seven of eight exemplar structures, the main $\mathrm{time}_b$ effect on $\mu$ was significant, indicating detectable within-subject change over follow-up. 
The direction and magnitude of change varied by organ and often depended on baseline age: for several soft-tissue organs (e.g., liver, kidney, spleen, gluteus, and heart), the model indicated predominantly declining trajectories across adulthood, whereas aorta, lung, and vertebrae increased over time (Fig.~\ref{fig:long_vol}).
Age-by-time interactions further showed that longitudinal slopes are not constant: for example, aortic growth was steeper at younger baseline ages and attenuated with increasing $\mathrm{age}_b$, whereas decreases in liver, kidney, and gluteus minimus became more pronounced at older baseline ages. 
Sex-by-time interactions were significant for spleen, liver, and gluteus minimus, indicating smaller proportional longitudinal change in men than in women, i.e., flatter longitudinal slopes in men.

\paragraph{Longitudinal attenuation (GAMM).}
Longitudinal attenuation was analyzed using generalized additive mixed models (GAMM). 
A prominent pattern in Fig.~\ref{fig:long_int} is contrast-dependent divergence in longitudinal trends: for liver, kidney, spleen, and aorta, attenuation tended to decrease over time in non-contrast scans but increase in contrast-enhanced scans. This reversal was strongest for contrast-enhanced kidney (estimated change 2.20 HU/year) and spleen (1.13 HU/year).
In non-contrast liver, attenuation decreased over time, with a strong baseline-age dependence (time-by-age interaction $\beta_{\mathrm{time}_b:\mathrm{age}_b}=0.24$, Bonferroni-significant), implying that the negative slope progressively flattens with higher baseline age. 
Beyond this reversal, the strongest baseline-age interaction was observed for the contrast-enhanced aorta ($\beta_{\mathrm{time}_b:\mathrm{age}_b}=0.44$), implying a significantly steeper increase at older baseline ages.

A second interaction-dominated pattern was also evident for gluteus minimus right: time effects were significant in both contrast states (0.92 non-contrast; 0.64 contrast), but both showed significant negative time-by-age interactions, such that the net longitudinal slope decreased with baseline age and was negative for essentially all adult ages; in addition, contrast-enhanced gluteus minimus showed a significant sex modification of the slope ($\beta_{\mathrm{time}_b:\mathrm{sex}}=0.20$).
Sex-related modification was otherwise uncommon, except for vertebrae L3 in contrast-enhanced scans.

\begin{table*}[!tbp]
\footnotesize
\caption{\label{tab:long_models}
Longitudinal fixed-effect estimates for the eight exemplar structures with follow-up time $\mathrm{time}_b$ (in years) and age at baseline $\mathrm{age}_b$. 
\textbf{Volume:} mixed-effects GAMLSS--GG for $\mu$ (log link); entries are shown as $\exp(\hat\beta)$ (multiplicative change in $\mu$ per 1-year increase in $\mathrm{time}_b$; interactions modify the longitudinal slope by baseline age and sex).
\textbf{Attenuation:} GAMM4 (identity link), fit separately for non-contrast ($\times$) and contrast-enhanced ($\checkmark$) scans; entries are additive changes in HU per 1-year increase in $\mathrm{time}_b$. Baseline age term uses $\mathrm{age}_b/10$.
Boldface marks Bonferroni-adjusted $p<0.05$. Coefficients for all structures in Supplement S12--13.}
\centering
\begin{tabular}{p{4cm}crrr}
\toprule
\textbf{Volume} &  & $\mathrm{time}_b$ & $\mathrm{time}_b\!:\!\mathrm{age}_b$ & $\mathrm{time}_b\!:\!\mathrm{sex}_\mathrm{M}$ \\
\midrule
aorta &  & \bftab 1.0307 & \bftab 0.9965 & 1.0000 \\
gluteus minimus right &  & \bftab 1.0035 & \bftab 0.9985 & \bftab 0.9985 \\
heart &  & \bftab 0.9932 & 1.0006 & 1.0013 \\
kidney right &  & \bftab 1.0137 & \bftab 0.9950 & 0.9986 \\
liver &  & \bftab 1.0080 & \bftab 0.9971 & \bftab 0.9956 \\
lung upper lobe right &  & \bftab 1.0118 & \bftab 0.9982 & 0.9996 \\
spleen &  & 1.0022 & \bftab 0.9992 & \bftab 0.9940 \\
vertebrae L3 &  & 1.0012 & 1.0002 & 1.0001 \\
\midrule
\textbf{Attenuation} & Contrast & $\mathrm{time}_b$ & $\mathrm{time}_b\!:\!\mathrm{age}_b$ & $\mathrm{time}_b\!:\!\mathrm{sex}_\mathrm{M}$ \\
\midrule
aorta & $\times$ & -0.6994 & 0.0812 & -0.0330 \\
aorta & $\checkmark$ & -0.2304 & \bftab 0.4402 & 0.4245 \\
gluteus minimus r & $\times$ & \bftab 0.9243 & \bftab -0.3857 & 0.2073 \\
gluteus minimus r & $\checkmark$ & \bftab 0.6393 & \bftab -0.3701 & \bftab 0.2007 \\
heart & $\times$ & 0.0131 & -0.0077 & -0.0895 \\
heart & $\checkmark$ & 0.6890 & 0.1562 & 0.3690 \\
kidney right & $\times$ & -0.7540 & -0.0233 & -0.0690 \\
kidney right & $\checkmark$ & \bftab 2.1986 & -0.0965 & 0.6538 \\
liver & $\times$ & \bftab -2.1696 & \bftab 0.2372 & 0.0548 \\
liver & $\checkmark$ & 1.2836 & -0.0287 & 0.1422 \\
lung upper lobe r & $\times$ & -1.9524 & 0.1189 & -0.2383 \\
lung upper lobe r & $\checkmark$ & -2.1170 & 0.2585 & -0.1668 \\
spleen & $\times$ & -0.0177 & -0.0476 & -0.1173 \\
spleen & $\checkmark$ & \bftab 1.1313 & 0.0382 & 0.1776 \\
vertebrae L3 & $\times$ & -0.7631 & -0.8162 & 1.9235 \\
vertebrae L3 & $\checkmark$ & -0.9732 & -0.3075 & \bftab 1.6543 \\
\bottomrule
\end{tabular}
\end{table*}

\newpage

\section{Discussion}

By aggregating over 350,000 CT examinations from five cohorts and applying automated segmentation of 106 anatomical structures, this study presents the most comprehensive reference framework for quantitative whole-body CT biomarkers across adulthood to date. 
The evidence-grounded, cross-verified LLM filtering pipeline addresses a central barrier to CT reference modeling: reliance on pathology-enriched clinical data due to the impracticality of population-scale whole-body CT in healthy volunteers. 
Combining report-based pathology filtering with distributional modeling yields reference distributions for organ volume and structure-wise CT attenuation. 
These cross-sectional and longitudinal models translate raw measurements into individualized, covariate-adjusted centile scores for research and opportunistic screening.

\paragraph{Pathology filtering from reports enables scalable CT reference modeling.}
To limit pathology leakage that would bias CT reference distributions, we developed a radiology report filtering pipeline for cohort construction. Five open-weight LLMs propose structure-level abnormality candidates with verbatim report support (Stage~1), and disagreements are adjudicated in an independent verification stage using pooled evidence (Stage~2).
Reliability is supported by three design elements: canonical structure identifiers, schema-constrained outputs, and explicit verbatim evidence from reports. 

Cross-verification reduced model dependence, improved agreement with manual annotations, and supported high-recall filtering to minimize pathology leakage that would bias reference trajectories and centiles. 
Unlike majority voting, evidence-based adjudication resolves disagreements using the report content rather than aggregating model outputs. 
Performance varied across datasets, indicating that achievable accuracy depends on report characteristics (e.g., language and style) as well as model choice and motivating site- or language-specific evaluation in new deployments. 
{The supplementary region-stratified analysis further indicates that filtering performance also varies anatomically, underscoring the need for region-aware validation when transferring the pipeline to new reporting settings.}
{Importantly, report-based filtering has an inherent ceiling: abnormalities not described in the clinical report cannot be removed by the LLM pipeline, irrespective of extraction accuracy.}
Based on the weighted aggregate, MedGemma Stage~2 resulted in the best overall performance and was used for downstream cohort filtering.

\paragraph{Pathology filtering changes reference distributions and centile utility.}
A key implication of report-based pathology reduction is that it alters the inferred reference distribution. Our comparisons of filtered and non-filtered attenuation models demonstrated that filtering can shift central trajectories and reshape distribution tails. This is methodologically consequential: tail behavior governs extreme centiles and is therefore central for individualized scoring and for screening-oriented use cases that rely on detecting atypical deviations. Consistent with this, filtered reference models yielded improved case--control discrimination compared with non-filtered references. 
In downstream centile scoring, we observed clear disease-associated shifts: cardiomegaly cases concentrated in the upper tail of heart-volume centiles, and reports with lung parenchyma findings showed significantly elevated lung-lobe attenuation centiles. 
Because the underlying subjects were held fixed and only the reference model changed, the observed AUC gains when using filtered versus non-filtered attenuation references directly quantify the benefit of pathology reduction for centile-based discrimination.
{At the same time, the modeled endpoints in this study are aggregate structure-level biomarkers, i.e., whole-structure volume and mean HU. This makes the charts relatively robust to small focal incidental findings, while diffuse or rare residual pathology may still affect the distributions, particularly the outer centiles.}
Together, these results show that pathology filtering is essential for meaningful centile interpretation in clinical CT cohorts and that evidence-grounded report parsing provides a scalable mechanism to achieve it.

\paragraph{Updated organ-volume reference charts.}
In parallel to the attenuation reference charts, the same pathology-reduced, multi-cohort corpus enables a substantive update of our previously published CT organ-volume charts \citep{wachingerBodyChartsCT2025}.
Relative to that work, we incorporated additional cohorts (including INSPECT and Merlin), applied report-based pathology filtering, and performed the longitudinal analysis with GAMLSS. These updates strengthen the interpretability of volumetric centiles by reducing pathology-driven tail inflation and by providing a coherent reference for cross-sectional and longitudinal trajectories.
Volumetric centiles quantify atypicality relative to an age- and covariate-matched reference, but for structures where absolute size is clinically actionable (e.g., aortic caliber), they should complement rather than replace established thresholds.

\paragraph{HU reference charts require distribution- and protocol-aware modeling.}
The HU reference charts provide a distributional characterization of tissue composition across adulthood at a whole-body anatomical scale. Across structures, attenuation distributions were frequently asymmetric and heavy-tailed, violating Gaussian assumptions that underpin many prior single-organ studies and simple reference-range approaches. The four-parameter skew-$t$ type-1 family captured both skewness and tail behavior, enabling coherent centile estimation across organs with qualitatively different distributional shapes. 
Extensive bootstrap analyses and diagnostic plots confirmed the reliability of the estimated GAMLSS-ST1 models, supporting it as a default for CT attenuation modeling. 

The fitted charts further demonstrate that both the expected attenuation ($\mu$) and the between-person dispersion ($\sigma$) vary nonlinearly with age across most structures. 
Modeling $\sigma$ goes beyond mean-focused HU analyses of prior studies \citep{boutinInfluenceIVContrast2016,holcombeVariationAortaAttenuation2022} by quantifying how  attenuation variability evolves across adulthood.
Fractional polynomials selected via BIC provided a compact representation that nevertheless captured complex patterns in central tendency and heteroscedasticity. The resulting FP specifications can serve as informative starting points for future studies with smaller sample sizes by reducing model-search burden and the risk of overfitting, while preserving the ability to capture nonlinear aging trajectories.
{Because the measurements are derived from automated segmentations, segmentation uncertainty can affect the estimated charts. Random segmentation variability would primarily add measurement noise and broaden the estimated distributions, whereas systematic segmentation bias could shift the expected trajectories. Robust outlier removal and inter-organ consistency checks were therefore used to reduce the influence of gross segmentation failures, although residual structure-specific segmentation uncertainty may remain.}

Acquisition factors were also prominent, reinforcing that HU reference modeling must explicitly account for protocol variability. Contrast enhancement induces large tissue- and compartment-specific shifts, motivating separate models for contrast-enhanced and non-contrast examinations. Tube potential and manufacturer effects were frequently detectable, consistent with known attenuation dependencies \cite{zhengTubeVoltageHUCorrection2020,aapmReport233CTPerformance}. These findings underscore that clinical deployment of HU-based centiles benefits from covariate-aware scoring that conditions on acquisition parameters. 

\paragraph{HU trajectories align with tissue-composition biology.}
Across tissue classes, the inferred age trends, sex differences, and acquisition effects align with established interpretations of CT attenuation as a marker of tissue composition, supporting the biological plausibility of the estimated centiles. In parenchymal organs, decreasing non-contrast attenuation with age is consistent with progressive fat infiltration \cite{meierAssessmentAgeRelatedChanges2007,hahnLongitudinalChangesLiver2015} and related observations in pancreas \cite{bhallaAssociationPancreaticFatty2022}. In skeletal muscle, attenuation decline with age reflects increasing myosteatosis with systematic sex differences \cite{figueiredoComputedTomographybasedSkeletal2021,graffyDeepLearningbasedMuscle2019}. In bone and cartilage, attenuation patterns across vertebrae/ribs and costal cartilage are compatible with osteoporosis-related decline and age-associated calcification, and the observed contrast-dependent shifts motivate protocol-aware interpretation \cite{boutinInfluenceIVContrast2016}. In vascular structures, contrast-enhanced aortic attenuation exhibits sex differences that are absent in non-contrast settings, consistent with prior reports emphasizing contrast state as a key determinant of vascular HU differences \cite{holcombeVariationAortaAttenuation2022}. Together, these consistencies support interpreting HU centiles as structure- and protocol-conditioned summaries of tissue composition across adulthood. 

\paragraph{Longitudinal change differs from cross-sectional aging.}
A unique advantage of the TUM PACS cohort is repeated imaging, enabling within-subject analyses that use individuals as their own control and reduce between-person variability. Within-subject HU change often diverged in both direction and magnitude from cross-sectional age associations, suggesting that cross-sectional trends may not reflect true within-person change in routine-care cohorts. 
This divergence likely reflects a mixture of biological change and cross-sectional composition effects like cohort effects and survivor bias. 
For organ volumes, we used mixed-effects GAMLSS to extend the cross-sectional models and quantify within-subject change in a coherent distributional framework. For attenuation, convergence limitations of highly parameterized longitudinal ST1 models prompted GAMM-based mean modeling as a pragmatic compromise. Developing scalable longitudinal distributional HU models that retain explicit tail and skewness parameterization remains an important methodological direction.

\paragraph{Comparison to prior longitudinal attenuation studies.}
Our longitudinal HU analysis uses a more flexible, covariate-aware modeling framework than prior studies. When we instantiate our models at the ages and acquisition settings reported in these studies, we can directly compare implied slopes to published estimates. For non-contrast liver, the literature reports $-0.35$~HU/year \cite{hahnLongitudinalChangesLiver2015}, whereas our model evaluated at the same setting yields a steeper decline of $\approx-0.8$~HU/year (female $-0.82$~HU/year; male $-0.77$~HU/year). For non-contrast L3 skeletal muscle, prior work reports $-1.45$~HU/year \cite{graffyDeepLearningbasedMuscle2019}, while our iliopsoas-based analogue (a narrower target than an aggregate L3 muscle compartment) yields $-0.51$ (female) to $-0.78$ (male)~HU/year and is further attenuated in contrast-enhanced examinations \cite{boutinInfluenceIVContrast2016}. For vertebral trabecular bone, published rates of $-2.5$~HU/year \cite{jangOpportunisticOsteoporosisScreening2019} closely match the mean slope obtained by averaging our contrast- and sex-stratified L1 estimates at the same reference age ($\approx-2.4$~HU/year), see Supplementary Note~S14. Together, these comparisons contextualize our longitudinal estimates relative to prior cohorts and protocols and support the plausibility of the inferred trajectories.

{Although the dataset is multi-source, a large proportion of examinations originated from the TUM PACS. This may introduce institutional bias related to reporting style, patient population, referral patterns, clinical indications, scanner/protocol distributions, and local acquisition practices. Our study-effect sensitivity analysis showed that typical source-study deviations from the pooled reference were modest for most structures, and TUM showed small deviations for volume models. However, these analyses cannot fully substitute for external validation in additional healthcare systems, particularly in cohorts with different demographic composition, reporting conventions, and acquisition protocols.}

{A related challenge is the limited availability of anthropometric variables in retrospective PACS and public CT datasets. Height, weight, and BMI may improve modeling of body-size-dependent volumes and body-composition-sensitive attenuation values. However, anthropometric adjustment requires careful consideration: because body composition can reflect the biological or pathological processes that centile scoring aims to capture, adjustment for weight or BMI may also mask clinically relevant deviations. Future work should therefore evaluate anthropometrically adjusted charts in cohorts where these variables are consistently available.}

\paragraph{Limitations.}
Several limitations warrant consideration. 
First, the reference charts begin in adulthood; pediatric reference curves remain to be developed. 
{Second, although the cohort spans multiple source studies, countries, vendors, and acquisition settings, it is derived from routine clinical CT and is not population-representative.}
{The study-effect analyses indicated modest source-study deviations for most structures, but broader validation across additional healthcare systems remains necessary, particularly with respect to demographic composition, clinical indications, protocol distributions, and reporting conventions.}
{Age was explicitly modeled, but age support was not uniform across adulthood; estimates in sparsely represented age ranges should therefore be interpreted together with the bootstrap uncertainty intervals.}
{Individual-level ethnicity, height, weight, and BMI were not consistently available and could not be included.}
{Third, report-based filtering yields a pathology-reduced rather than pathology-free cohort; unreported abnormalities cannot be removed by the LLM pipeline, and residual pathology may affect the outer centiles, which should therefore be interpreted as empirical reference limits rather than strict thresholds of normality.}
{Fourth, the measurements depend on automated segmentation; although prior validation, robust outlier removal, and inter-organ consistency checks mitigate gross failures, residual segmentation uncertainty may affect the derived centiles.}
Finally, radiation exposure is an inherent limitation of CT-based screening; however, the intended use of these charts is opportunistic, so applying the charts does not incur additional dose.
{Extending report filtering to other languages or reporting environments will require validation in the target setting.}

\paragraph{Conclusion.}
In summary, structure-wise reference charts translate raw whole-body CT biomarkers into interpretable, covariate-adjusted centiles and provide a shared coordinate system for quantitative imaging across adulthood.
By coupling distribution-aware modeling with LLM-based pathology filtering, this work enables reference modeling from routine clinical CT at a scale and anatomical breadth that was previously difficult to achieve.
{These resources advance opportunistic screening research and standardized quantitative phenotyping across sites, and motivate PACS-integrated, covariate-adjusted centile scoring for routine CT. In such a workflow, automated segmentation and scoring could run in the background after image acquisition and provide structure-wise volume and attenuation centiles during reporting, complementing qualitative assessments with age-, sex-, and acquisition-adjusted quantitative context. These outputs are intended as decision support rather than standalone diagnostic labels and require prospective validation before routine clinical use.}
The interactive whole-body charts are available at \url{http://www.ai-med.de/body-charts-v2}.

\newpage
\section{Methods}

\subsection{Participants and Cohort Construction}
\label{sec:cohort}
This retrospective study was approved by the institutional ethics committee (Approval No. 2025-99-S-CB). We compiled CT examinations from five sources: our institutional research PACS (TUM), the TotalSegmentator CT dataset \cite{wasserthalTotalSegmentatorRobustSegmentation2023}, CT-Rate \cite{hamamciDevelopingGeneralistFoundation2025}, INSPECT \cite{huangINSPECTMultimodalDataset2023}, and Merlin \cite{blankemeierMerlinVisionLanguage2024}. The pooled sample comprised adults aged 18--98 years scanned on CT systems from Philips, Siemens, GE, and Toshiba. For TUM, CT examinations acquired between September 2012 and August 2025 were retrieved from a research PACS storing pseudonymized metadata. DICOM data were obtained via query--retrieve and converted to NIfTI for downstream processing.

The included sources span different clinical scopes: CT-Rate provides non-contrast chest CT, INSPECT provides contrast-enhanced CT pulmonary angiography, and Merlin provides abdominal CT with predominantly contrast-enhanced acquisitions, whereas TUM and TotalSegmentator include mixed body regions and both non-contrast and contrast-enhanced studies (dataset characteristics in Supplement S2.3). The five sources together cover multiple sites in Germany, Switzerland, Turkey, and the United States.

Cohort construction proceeded in three stages. First, 351{,}915 CT examinations were processed with automated segmentation, yielding 12{,}966{,}897 structure segmentations across 106 anatomical targets. These comprised TUM (292{,}914 examinations), CT-Rate (19{,}399), INSPECT (13{,}677), Merlin (24{,}716), and TotalSegmentator (1{,}209). 
Second, two-stage outlier removal based on segmentation-derived volume statistics (see Supplement S2.1) reduced the dataset to 308{,}022 examinations from 194{,}936 participants (12{,}687{,}719 segmentations). Third, report-based pathology filtering (see Sec. \ref{sec:llmFilt}) further reduced the reference cohort to 276{,}132 examinations from 177{,}014 participants, totaling 9{,}179{,}609 structure segmentations (Supplementary Table~S1).

\subsection{Radiology Report Filtering via a Cross-Verified LLM Ensemble}
\label{sec:llmFilt}
To construct pathology-reduced cohorts for reference modeling, radiology reports were used to identify anatomical structures affected by clinically relevant abnormalities. Because routine clinical datasets do not provide structured, structure-level pathology labels at the scale required for reference chart construction, we implemented an automated filtering pipeline based on a cross-verified ensemble of open-weight large language models (LLMs). 
The aim was not diagnostic classification, but exclusion of potentially pathological structures to reduce pathology leakage into the reference sample. The pipeline couples schema-constrained, evidence-grounded extraction of candidate abnormalities with cross-verification of disputed findings, thereby separating candidate generation from adjudication and enabling transparent, auditable decisions at scale.
Prompts and implementation details are provided in Supplementary Section~S\textit{2.2}.

All segmentations were mapped to 39 canonical anatomical target classes (KANON) by merging left/right counterparts and consolidating repetitive structures (e.g., ribs and vertebrae). 
Each class was assigned a unique structure identifier, which served as a stable reference space for all models and both report languages. This standardization was essential for reliable aggregation: an initial implementation that allowed models to output free-form structure names led to hallucinated structures and minor wording variants that prevented consistent voting and downstream processing. Using canonical identifiers as a shared ``lingua franca'' resolved these issues and enabled deterministic parsing, comparison, and auditing across models and datasets.

\paragraph{Model selection and scope.}
We selected five open-weight LLMs to maximize robustness of cross-verification across \emph{languages}, \emph{model lineages}, and \emph{domain specialization}. Qwen3-32B and Qwen2.5-72B were chosen as strong multilingual instruction-tuned generalists, matching our English/German report mix and providing an internal contrast in model scale within the same lineage~\citep{qwen3_report,qwen25_report}. Llama~3.3-70B was included as a high-capacity model from an independent  lineage, increasing ensemble diversity and reducing the likelihood of correlated errors driven by a single training and alignment recipe~\citep{llama33_modelcard,llama3_herd}. To test whether medical domain adaptation improves evidence-based adjudication of abnormalities, we incorporated OpenBioLLM-70B (biomedical adaptation of Llama) and MedGemma-27B (text-only medical model based on Gemma)~\citep{openbiollm70b_modelcard,medgemma_modelcard_v1,medgemma_techreport}. Reports were processed in their native language (English or German) without translation.

\paragraph{Cross-verified, evidence-grounded LLM filter.}
In Stage~1, each model was instructed to flag only abnormal KANON targets and to ground each abnormality by copying a supporting sentence verbatim from the report. Outputs were constrained via guided decoding to a strict JSON schema, with one record per flagged structure containing the structure identifier, canonical structure name, report-derived structure name, and the verbatim evidence sentence. Decoding was deterministic (\texttt{temperature}=0, \texttt{top\_p}=1). 
This design grounded every abnormality claim in explicit report text and made the extraction step machine-actionable and auditable. Although models were instructed to output only abnormal structures, occasional normal entries were produced and were removed during post-processing. 
Structures classified as abnormal by all five models were excluded without further adjudication.

In Stage~2, non-unanimous findings were adjudicated by evidence-only cross-verification. For each disputed structure (flagged by at least one but not all models), we pooled and deduplicated the evidence sentences from Stage~1, removed model identities, and presented only the structure identifier, canonical target name, and pooled evidences to the verifier. The full report text was intentionally withheld to constrain decisions to explicitly cited evidence and reduce over-interpretation. 
Verification applied a strict two-part criterion: (i) the cited sentence had to unambiguously refer to the same anatomy as the canonical structure, and (ii) it had to state a pathological or non-physiologic finding.

\paragraph{Decision strategy and robustness assessment.}
Each LLM produced an abnormality set after Stage~1 and after Stage~2. As aggregation baselines, we considered majority voting across the five models applied to the Stage~1 outputs and to the Stage~2 outputs. Majority voting aggregates model predictions at the level of structure identifiers and does not consider the underlying report evidence. In contrast, Stage~2 adjudication conditions explicitly on the pooled evidence sentences extracted from the reports, enabling disagreement resolution driven by cited text rather than by vote counts alone. In the operational pipeline, final abnormality labels were taken from the Stage~2 output of MedGemma-27B.

{Robustness was assessed using inter-model agreement metrics (pairwise Jaccard overlap and the proportion of cases with identical structure sets across all models) and by comparison to manual structure-level annotations on a subset of reports (TUM: $n=198$, Merlin: $n=100$, CT-Rate: $n=81$, INSPECT: $n=78$). Manual annotations were performed at the level of the 39 canonical anatomical targets by the first author, who identified all report-described abnormal target structures from the report text. To assess anatomical variation in filtering performance, we additionally grouped the 39 canonical targets into broader anatomical regions and computed region-stratified Jaccard overlap, precision, and recall.}
To minimise contamination of the reference distributions, any structure flagged as abnormal was excluded from statistical modeling.
Report-based filtering was applied to TUM, CT-Rate, INSPECT, and Merlin, which provide matched radiology reports. For the TotalSegmentator dataset, scans were excluded based on pathology indicators provided in the dataset metadata.

\subsection{Image Analysis}
\label{sec:image}
We used TotalSegmentator version 2.2.1 \cite{wasserthalTotalSegmentatorRobustSegmentation2023} for the automated segmentation of 106 anatomical structures, including 23 organs, 58 bones, 8 muscles, and 17 major vessels (see supplement S1); segmentation quality on our PACS data was previously validated~\citep{wachingerBodyChartsCT2025}. 
Mean HU values were then extracted for every structure. 
We used a two-stage outlier removal strategy anchored to the manually reviewed TS dataset \citep{wachingerBodyChartsCT2025}. First, we removed extreme structure volumes using robust median/MAD-based filtering on log-transformed volumes. Second, we flagged additional outliers by identifying scans that violated learned inter-organ dependencies using low-rank matrix factorization (Supplement~S2.1).

\subsection{Cross-sectional GAMLSS\label{sec:methodCrossSec}}
We constructed reference distributions separately for each anatomical structure, treating the structure-level summary statistic from each CT examination as the unit of analysis (mean attenuation in HU or structure volume). 
For cross-sectional reference charts, we used generalized additive models for location, scale, and shape (GAMLSS)~\cite{rigbyGeneralizedAdditiveModels2005}, the framework endorsed by the WHO for reference chart construction~\cite{borghiConstructionWorldHealth2006}. GAMLSS are distributional regression models, i.e., they do not only estimate the mean (location $\mu$) but the full distributional characteristics of the outcome $Y$, including scale $\sigma$, skewness $\nu$, and kurtosis/tail-weight $\tau$. All aspects of the distribution can be modeled with non-linear functions of the explanatory variables. 

{Intuitively, this framework is useful for reference charts because it estimates the complete covariate-adjusted distribution of a measurement, rather than only its average value. Fractional polynomials allow the expected trajectory across age to bend smoothly without imposing a strictly linear trend, while avoiding the complexity of highly flexible spline models. Distributional regression further allows the spread and shape of the distribution to vary with covariates, which is important because centile scores depend on where a measurement lies within the full reference distribution, including its tails.}

Formally, the GAMLSS framework is specified as:
\begin{align}
Y &\sim \mathcal{D}\!\left(\mu,\sigma,\nu,\tau\right) \\
g_{\mu}\!\left(\mu\right) &= X_{\mu}\beta_{\mu} + Z_{\mu}\gamma_{\mu} \\
g_{\sigma}\!\left(\sigma\right) &= X_{\sigma}\beta_{\sigma} + Z_{\sigma}\gamma_{\sigma} \\
g_{\nu}\!\left(\nu\right) &= X_{\nu}\beta_{\nu} + Z_{\nu}\gamma_{\nu} \\
g_{\tau}\!\left(\tau\right) &= X_{\tau}\beta_{\tau} + Z_{\tau}\gamma_{\tau}
\end{align}
with the probability distribution $\mathcal{D}$, link functions $g(\cdot)$, fixed effects coefficients $\beta$ with corresponding design matrices $X$, and random effects coefficients $\gamma$ with design matrices $Z$. For modeling non-linear age effects, we used fractional polynomials (FP)~\cite{roystonRegressionUsingFractional1994}, offering flexibility without excess complexity. Fractional polynomials enter the model via the $X$ terms with associated coefficients $\beta$.

We modeled CT attenuation with the Azzalini skew-$t$ type-1 (ST1) family: a four-parameter generalization of the Gaussian with explicit skewness ($\nu$) and tail-weight ($\tau$) terms~\cite{azzaliniSkewnormalDistribution1986}. With identity links for $\mu$ and $\nu$, together with log links for $\sigma$ and $\tau$, ST1 handles real-valued HU data, including negative values, without transformation. The skewness $\nu$ flexibly captures the asymmetry observed in CT attenuation distributions, and $\tau$ provides robustness against heavy-tailed extremes, yielding a more accurate fit in the presence of outliers than a standard Gaussian. As imaging contrast has a wide impact on the HU values, we fitted separate models for scans with and without contrast agent. The final regression equation for each structure was:
\begin{align}
Y &\sim \mathrm{ST1}\!\left(\mu,\sigma,\nu,\tau\right) \\
\mu &= \beta_{\mu}
      + \beta_{\mu,\mathrm{sex}}\!\left(\mathrm{sex}\right)
      + \beta_{\mu,\mathrm{manu}}\!\left(\mathrm{manu}\right)
      + \beta_{\mu,\mathrm{kvp}}\!\left(\mathrm{kvp}\right)      
      + \mathrm{FP}_{\mu}\!\left(\mathrm{age}\right)
      + \gamma_{\mu,\mathrm{study}}, \\
\log(\sigma) &= \beta_{\sigma}
      + \beta_{\sigma,\mathrm{sex}}\!\left(\mathrm{sex}\right)
      + \mathrm{FP}_{\sigma}\!\left(\mathrm{age}\right)
      + \gamma_{\sigma,\mathrm{study}}      
      , \\
\nu &= \beta_{\nu}, \\
\log(\tau) &= \beta_{\tau}.
\end{align}
Manufacturer and tube voltage (kVp) were included in the location submodel to capture systematic acquisition-related shifts in expected attenuation. Age and sex were included in both the location and scale submodels to capture nonlinear biological trends and age-dependent heteroscedasticity. Study-specific random intercepts were included in both location $\gamma_{\mu,\mathrm{study}} \sim \mathcal{N}\!\left(0,\delta^2_\mu\right)$ and scale $\gamma_{\sigma,\mathrm{study}}  \sim \mathcal{N}\!\left(0,\delta^2_\sigma\right)$ to account for residual between-cohort differences not explained by measured covariates.
Age enters through the fractional polynomials for location $\mathrm{FP}_{\mu}\!\left(\mathrm{age}\right)$ and scale $\mathrm{FP}_{\sigma}\!\left(\mathrm{age}\right)$.
The skewness and tail-weight parameters were modeled as intercept-only terms to avoid overparameterization of already highly flexible ST1 models.
The GAMLSS fitting routine selects the optimal set of FP powers from the standard set $\{-2,-1,-0.5,0,0.5,1,2,3\}$, where each power represents an age term (with $\log(\mathrm{age})$ for $p=0$) in the design matrix. We permitted FP expansions up to degree~3, i.e., up to three power terms, for location and scale (see Supplement~S2.4 for details). The optimal fractional polynomials for $\mathrm{FP}_{\mu}$ and $\mathrm{FP}_{\sigma}$ were selected based on the Bayesian information criterion (BIC) and reported in Table \ref{tab:FP_all}, and Supplement S10--11 for all 106 structures.  
Because linear terms were included in the selection set, nonlinearity was not enforced a priori; it was selected only when supported by BIC.

For organ volumes, we built on our prior whole-body volume charting framework \citep{wachingerBodyChartsCT2025} and extend it by adding tube voltage as a covariate, increasing multi-study coverage, and re-estimating the models on the pathology-reduced cohort.
We used a generalized Gamma (GG) distribution,
$Y \sim \mathrm{GG}\!\left(\mu,\sigma,\nu\right)$.
We fitted a single model across contrast conditions and include contrast status as a covariate.
We used log links for $\mu$ and $\sigma$ and an identity link for $\nu$; $\nu$ was modeled as an intercept-only term.
Analogous to the attenuation models, we included study-specific random intercepts in the location and scale.

Goodness of fit was assessed with Q--Q plots and detrended transformed Owen plots (DTOPs), the latter enabling direct comparison across alternative distributions (Supplement~S2.6). To evaluate model reliability and stability, we performed bootstrap analysis with 1{,}000 resamples stratified by study and sex to preserve the relative proportions of the original data. Reference trajectories and 95\% confidence intervals were then computed from the bootstrap replicates, for $\mu$ and $\sigma$, indicating that central tendency and dispersion are robustly estimated.
{Finally, we assessed the potential impact of study effects by analyzing the study-specific random-effect terms in the GAMLSS framework. We extracted best linear unbiased predictions (BLUPs) of study-specific deviations from the pooled reference trajectories, because these directly quantify the magnitude and direction of systematic shifts across source cohorts after accounting for the modeled covariates. For volume models with log-linked $\mu$ and $\sigma$ parameters, BLUPs were expressed as percentage deviations from the pooled reference trajectory or dispersion parameter, respectively. For attenuation models, BLUPs of the identity-linked $\mu$ parameter were expressed in HU, while BLUPs of the log-linked $\sigma$ parameter were expressed as percentage deviations in dispersion.}

\subsection{Individualized centile scores}
Individualized centile scores benchmark each subject's attenuation or volume per structure against age-specific reference distributions, where extreme values indicate atypical attenuation or volume. For each structure, the raw HU value or volume $y$ is converted to a centile by evaluating the cumulative distribution function (CDF) of the fitted GAMLSS model, conditional on the subject's covariates:
\begin{equation}
\mathrm{Centile}(y)=100\times F_{\mathcal{D}}\!\left(
y \,\middle|\, \mathrm{age},\, \mathrm{sex},\, \mathrm{manu},\, \mathrm{kvp},\, \mathrm{contrast}
\right),
\end{equation}
where $F_{\mathcal{D}}$ denotes the CDF of the distributional family $\mathcal{D}$ with subject-specific predicted parameters (ST1: $\hat{\mu},\hat{\sigma},\hat{\nu},\hat{\tau}$; GG: $\hat{\mu},\hat{\sigma},\hat{\nu}$).
For attenuation, centiles were computed using the contrast-specific HU model (separate models for contrast-enhanced and non-contrast examinations). For volumes, centiles were computed from a unified model that included contrast status as a covariate.
The resulting centile represents the proportion of the reference population expected to have a value at or below the observed measurement under the same covariate setting.

\subsection{Longitudinal models\label{sec:long}}
Longitudinal analyses were restricted to our PACS (TUM) cohort, which is the only dataset that provides repeated examinations with consistent linkage; Supplement~S2.3 overviews key characteristics of the sample (examinations per participant, follow-up span, and inter-scan intervals). 
Age was decomposed~\cite{wachingerLongitudinalImagingGenetics2018} as
$\mathrm{age} = \mathrm{age}_b + \mathrm{time}_b$,
where $\mathrm{age}_b$ is baseline age (age at the first examination) and
$\mathrm{time}_b$ is time from baseline (years since the first examination).
This decomposition allows non-linear baseline age effects to be modeled using the same FP specification as in the cross-sectional reference models, while $\mathrm{time}_b$ captures longitudinal change.

\paragraph{Longitudinal organ volumes.}
For organ volumes, longitudinal analyses were performed within the GAMLSS framework using the generalized Gamma distribution, building directly on the previously introduced cross-sectional models. The same FP specification for baseline age was reused for both the location and scale. A single model was fitted across contrast conditions, with contrast included as a covariate when available. The location predictor was:
\begin{align}
\log(\mu) &=
\beta_{0}
+ \beta_{\mathrm{sex}}\!\left(\mathrm{sex}\right)
+ \beta_{\mathrm{manu}}\!\left(\mathrm{manu}\right)
+ \beta_{\mathrm{kvp}}\!\left(\mathrm{kvp}\right)
+ \beta_{\mathrm{contrast}}\!\left(\mathrm{contrast}\right)
+ \mathrm{FP}_{\mu}\!\left(\mathrm{age}_{b}\right) \nonumber \\
&+ \beta_{\mathrm{time}_b}\!\left(\mathrm{time}_{b}\right)
+ \beta_{\mathrm{time}_b:\mathrm{age}_b}\!\left(\mathrm{time}_{b}:\mathrm{age}_{b}\right)
+ \beta_{\mathrm{time}_b:\mathrm{sex}}\!\left(\mathrm{time}_{b}:\mathrm{sex}\right)
+ \gamma_{\mathrm{ID}}.
\end{align}
$\gamma_{\mathrm{ID}}\sim\mathcal{N}(0,\delta^2)$ is a subject-specific random intercept, and $\mathrm{manu}$ and $\mathrm{kvp}$ are scan-level covariates.
Interaction terms allow longitudinal change to vary by baseline age and by sex.
The scale parameter retained the cross-sectional specification with chronological age replaced by baseline age. 
The shape parameter $\nu$ was initialized from the cross-sectional estimate and held fixed to improve stability.

\paragraph{Longitudinal HU (attenuation).}
For CT attenuation, longitudinal extensions of the cross-sectional ST1 model exhibited convergence problems for many structures, likely due to the complexity of the four-parameter ST1 family. 
We therefore fit a generalized additive mixed model (GAMM)~\cite{woodGeneralizedAdditiveModels2017} to estimate the conditional mean. Distributional modeling is essential for cross-sectional centile scoring and tail behavior, whereas the longitudinal analysis targets stable estimation of within-subject mean change.
We retain the FP-based baseline age specification from the cross-sectional HU GAMLSS model. Models were fitted separately for contrast-enhanced and non-contrast examinations. 
The location predictor was:
\begin{align}
\mu &=
\beta_{0}
+ \beta_{\mathrm{sex}}\!\left(\mathrm{sex}\right)
+ \beta_{\mathrm{manu}}\!\left(\mathrm{manu}\right)
+ \beta_{\mathrm{kvp}}\!\left(\mathrm{kvp}\right)
+ \mathrm{FP}_{\mu}\!\left(\mathrm{age}_{b}\right)
+ \beta_{\mathrm{time}_b}\!\left(\mathrm{time}_{b}\right) \nonumber \\
&\quad
+ \beta_{\mathrm{time}_b:\mathrm{age}_b}\!\left(\mathrm{time}_{b}:\mathrm{age}_{b}\right)
+ \beta_{\mathrm{time}_b:\mathrm{sex}}\!\left(\mathrm{time}_{b}:\mathrm{sex}\right)
+ \gamma_{\mathrm{ID}},
\end{align}
where $\mu$ denotes the conditional mean (identity link).

\section*{Supplement}
The supplementary matrial is available at: 
\url{https://github.com/ai-med/body-charts/blob/main/body_charts_supp.pdf}

\section*{Data availability}
Four of the five cohorts used in this study are publicly available. The TotalSegmentator CT dataset can be accessed via Zenodo at \url{https://doi.org/10.5281/zenodo.6802613}. The CT-Rate dataset is available at \url{https://huggingface.co/datasets/ibrahimhamamci/CT-RATE}. The INSPECT dataset is available via its project page, including access instructions, at \url{https://som-shahlab.github.io/inspect-website/}. The Merlin dataset is available via the official repository and download instructions at \url{https://github.com/StanfordMIMI/Merlin/blob/main/documentation/download.md}.
The in-house TUM cohort was extracted from a clinical research PACS and cannot be shared publicly due to data protection and institutional approvals.

\section*{Code availability}
Interactive body charts are available at \url{http://www.ai-med.de/body-charts-v2}.
Source code is available at \url{https://github.com/ai-med/body-charts}. 
Segmentation was performed with the publicly available TotalSegmentator (v2.2.1) \citep{wasserthalTotalSegmentatorRobustSegmentation2023}. 
LLM-based filtering used publicly released open-weight models obtained from Hugging Face in quantized form (AWQ weights with FP8 key–value caching): Qwen2.5-72B, Qwen3-32B, Llama 3.3-70B, OpenBioLLM-70B, and MedGemma-27B (text-only).

\bibliographystyle{unsrtnat}

\bibliography{MyLibrary}

\end{document}